\documentclass[10pt,twocolumn,letterpaper]{article}

\usepackage{cvpr}              

\usepackage{multirow}
\usepackage{makecell}
\usepackage{arydshln}
\usepackage{graphicx}
\usepackage{fontawesome}
\usepackage{bbding}
\usepackage{pifont}
\usepackage{cuted}
\usepackage{amsmath}
\usepackage{algorithm}
\usepackage{algpseudocode}
\usepackage{amsfonts}
\usepackage{amssymb}
\usepackage{cuted}
\usepackage{amssymb}
\usepackage{pifont}
\usepackage[accsupp]{axessibility}  

\definecolor{cvprblue}{rgb}{0.21,0.49,0.74}
\usepackage[pagebackref,breaklinks,colorlinks,allcolors=cvprblue]{hyperref}
\usepackage{caption}
\def\paperID{142} 
\def\confName{CVPR}
\def\confYear{2026}

\title{MorphAny3D: Unleashing the Power of Structured Latent in 3D Morphing}

\author{
    Xiaokun Sun$^{1}$, 
    Zeyu Cai$^{1}$,
    Hao Tang$^{2}$,
    Ying Tai$^{1}$,
    Jian Yang$^{1}$,
    Zhenyu Zhang$^{1*}$ \\
    $^{1}$Nanjing University \quad $^{2}$Peking University \\
    $^{*}$Corresponding Author \\
    {\tt\small xiaokun\_sun@smail.nju.edu.cn, caizeyu010612@gmail.com, bjdxtanghao@gmail.com} \\
    {\tt\small \{yingtai, csjyang\}@nju.edu.cn, zhangjesse@foxmail.com}
    }

\begin{document}

\twocolumn[{
\renewcommand\twocolumn[1][]{#1}
\maketitle
\vspace{-13mm}
\begin{center}
    \captionsetup{type=figure}
    \includegraphics[width=1.00\textwidth]{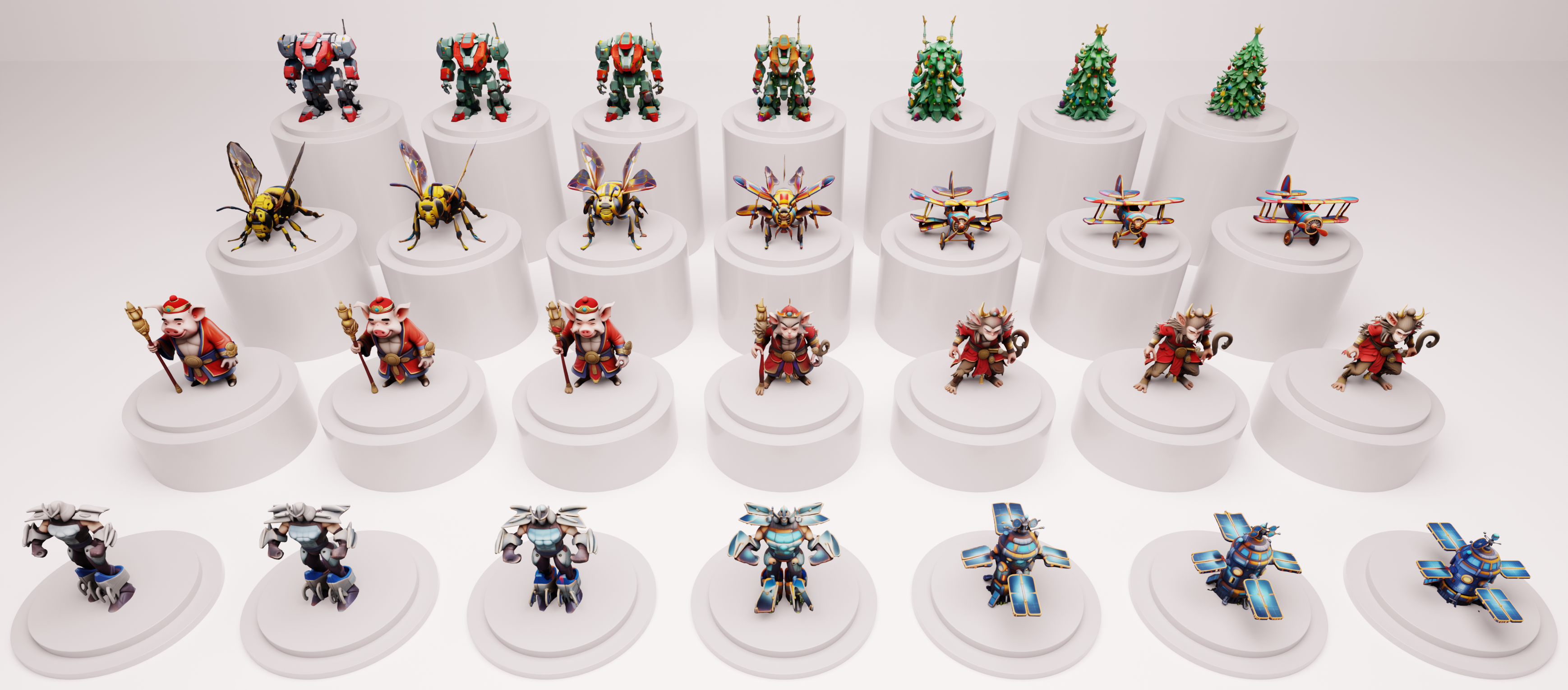}
    \caption{\textbf{MorphAny3D} enables smooth, semantically coherent, and aesthetically pleasing transitions from source objects (first column) to target objects (last column), even when they share no semantic or visual relationship (e.g., bee to biplane in the second row).}
    \label{fig:teaser}
\end{center}
}]

\begin{abstract}
3D morphing remains challenging due to the difficulty of generating semantically consistent and temporally smooth deformations, especially across categories. We present MorphAny3D, a training-free framework that leverages Structured Latent (SLAT) representations for high-quality 3D morphing. Our key insight is that intelligently blending source and target SLAT features within the attention mechanisms of 3D generators naturally produces plausible morphing sequences. To this end, we introduce Morphing Cross-Attention (MCA), which fuses source and target information for structural coherence, and Temporal-Fused Self-Attention (TFSA), which enhances temporal consistency by incorporating features from preceding frames. An orientation correction strategy further mitigates the pose ambiguity within the morphing steps. Extensive experiments show that our method generates state-of-the-art morphing sequences, even for challenging cross-category cases. MorphAny3D further supports advanced applications such as decoupled morphing and 3D style transfer, and can be generalized to other SLAT-based generative models.
Project page: \url{https://xiaokunsun.github.io/MorphAny3D.github.io/}.
\end{abstract}

\section{Introduction}
\label{sec:1}
Morphing~\cite{2Dsurvey1,2Dsurvey2,2D3Dsurvey} refers to the visual effect of seamlessly transforming a source into a target through a smooth, plausible, and aesthetic deformation sequence.
This foundational technique is widely used in animation, film, and game design to enhance creative expression.
Based on input dimensionality, morphing is categorized into 2D morphing~\cite{2Dmorphing1,2Ddatamorphing1,diffmorpher,freemorph} and 3D morphing~\cite{lmkmap1,datamap1,morphflow,gaussianmorphing,3Dmorpher}.
While 2D morphing has advanced significantly with diffusion models~\cite{diffusion_ddim,diffusion_ddpm,diffusion_sde,stable_diffusion}, 3D morphing remains challenging due to the inherent complexity of modeling smooth and reasonable deformations in three dimensions.

\begin{figure}[tp]
    \centering
    \includegraphics[width=0.45\textwidth]{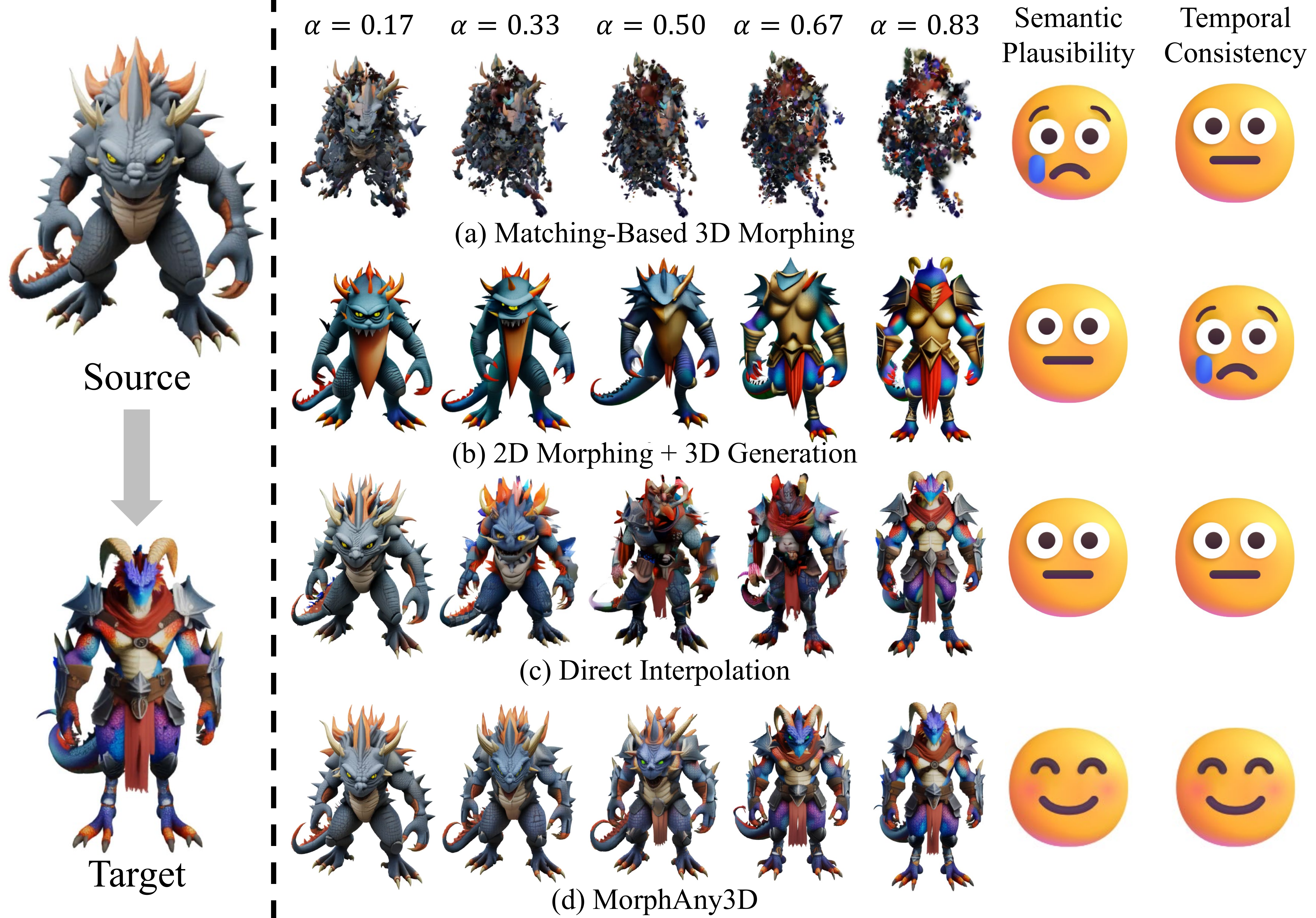} 
    \caption{Comparison of different 3D morphing strategies. (a) Matching-Based 3D Morphing; (b) 2D Morphing + 3D Generation; (c) Direct Interpolation; (d) MorphAny3D. Our method leverages the powerful SLAT to achieve semantically plausible and temporally smooth 3D morphing without any training. $\alpha \in [0, 1]$ is the deformation weight controlling the morphing progress.}
    \label{fig:comp}
    \vspace{-0.6cm}
\end{figure}

Most existing 3D morphing approaches follow a two-stage paradigm: (1) establishing dense correspondences~\cite{3Dmatch_survey1,3Dmatch_survey2} between source and target objects using hand-crafted sparse landmarks~\cite{lmkmap1,lmkmap2}, functional maps~\cite{functionmap3}, optimal transport~\cite{opt_trans1}, data priors~\cite{datamap1}, or extracted 2D features~\cite{featuremap2}; and (2) interpolating between these correspondences to generate intermediate shapes.
While effective in constrained settings, such matching-based methods face several critical limitations.
First, they typically prioritize geometric deformation while ignoring the concurrent evolution of textures.
Second, they often exhibit limited generalization, particularly in cross-category transformations (e.g., morphing a chair into a car).
As shown in \cref{fig:comp}-(a), inaccurate correspondence estimation frequently leads to structurally implausible morphing results.

Recent advances in feed-forward 3D generation frameworks~\cite{lgm,wonder3d,3DShape2VecSet,hunyuan3d_v2,clay,trellis,direct3d} have substantially improved the quality and fidelity of 3D synthesis.
Notably, Trellis~\cite{trellis} achieves diverse and high-fidelity 3D generation through its Structured Latent (SLAT) representation.
Owing to SLAT's explicit and regular structure, it shows strong potential for training-free applications such as 3D editing~\cite{voxhammer,nano3d}, stylization~\cite{stylesculptor}, and scene modeling~\cite{hiscene}.
A naive strategy for 3D morphing with SLAT involves first generating a 2D morphing sequence and then lifting each frame independently into 3D using Trellis.
However, as illustrated in \cref{fig:comp}-(b), this approach fails to ensure temporal consistency due to frame-wise generation and often misrepresents complex 3D deformations.
A strategy that is more deeply integrated with the generative framework is to directly interpolate the initial noise and conditional features of Trellis, similar to generative-prior-based morphing methods~\cite{diffmorpher,3Dmorpher}.
Yet, as shown in \cref{fig:comp}-(c), this strategy lacks explicit constraints on structural plausibility and temporal continuity, often producing suboptimal morphing quality.
In summary, while the SLAT representation offers compelling opportunities for 3D morphing, achieving smooth, high-fidelity, and temporally coherent 3D morphing within modern SLAT-based generative frameworks remains an open and pressing challenge.

To address these challenges, we present \textbf{MorphAny3D}, a novel and robust training-free 3D morphing framework built upon the SLAT representation.
As demonstrated in \cref{fig:teaser} and \cref{fig:comp}-(d), MorphAny3D fully utilizes the 3D generative prior embedded in SLAT to produce structurally plausible and temporally smooth morphing sequences between diverse object categories.
Our methodology stems from a key observation: rather than interpolating at the noise or condition level, directly aggregating SLAT features within attention mechanisms yields more reasonable and visually smooth 3D deformations.
Guided by this insight, we introduce two core attention-based components:
(1) Morphing Cross-Attention (MCA): intelligently fuses information from the source and target objects in the cross-attention layers, ensuring the structural coherence and aesthetics of the deformation.
(2) Temporal-Fused Self-Attention (TFSA): enhances temporal coherence by incorporating SLAT features from the previous morphing frame into the self-attention mechanism, enabling smooth transitions over time.
Furthermore, we propose an orientation correction strategy inspired by the orientation distribution patterns of SLAT-based 3D objects, which effectively resolves abrupt orientation shifts during the morphing process.
Extensive experiments demonstrate that our method generates more plausible, smoother, and aesthetically superior 3D morphing sequences compared to existing approaches.
Beyond basic morphing, MorphAny3D natively supports advanced applications such as decoupled morphing and 3D stylization, and it can be seamlessly transferred to other SLAT-based models.

Our main contributions are summarized as follows:
\begin{itemize}
    \item We propose MorphAny3D, the first training-free 3D morphing framework based on Structured Latent (SLAT), capable of generating smooth and semantically coherent deformations across diverse 3D object categories.
    \item Inspired by SLAT fusion patterns in attention mechanisms, we propose Morphing Cross-Attention and Temporal-Fused Self-Attention—two key components that leverage SLAT features across objects and frames to improve morphing quality and temporal consistency.
    \item Drawing on the statistical orientation distributions of SLAT-based 3D objects, we propose an orientation correction strategy to mitigate abrupt orientation changes, further enhancing morphing smoothness.
\end{itemize}

\section{Related Work}
\label{sec:2}
\textbf{2D Morphing.} 
Image morphing~\cite{2Dmorphing1,2Dmorphing2,2Dmorphing3,2Dmorphing4,2D3Dsurvey} has long been studied in computer vision and graphics due to its broad applications. 
Traditional methods~\cite{2Dmorphing1,2Dmorphing2,2Dmorphing3,2Dmorphing4} rely on hand-crafted features to establish image correspondences, followed by gradual blending to produce smooth transitions.
However, they often fail to generate plausible, novel content and are prone to artifacts such as ghosting.
With the advent of deep learning, data-driven approaches~\cite{2Ddatamorphing1,2Ddatamorphing2} have improved results by learning 2D generative priors. 
Nevertheless, their limited model capacity makes cross-category morphing challenging. 
The recent success of diffusion models~\cite{diffusion_ddim,diffusion_ddpm,diffusion_sde} has enabled high-fidelity and diverse image generation. 
By leveraging pre-trained text-to-image models~\cite{stable_diffusion}, several works~\cite{interpolating,Impus,aid,diffmorpher,freemorph} have significantly enhanced morphing quality across different categories.
Despite these advances, extending such smooth and semantically coherent transformations to 3D content remains an open challenge.

\noindent
\textbf{3D Morphing.} 
Most 3D morphing methods rely on correspondences between source and target shapes, followed by the interpolation of matched 3D primitives.
The core difficulty lies in accurate 3D matching~\cite{3Dmatch_survey1,3Dmatch_survey2,3Dmatch_survey3,3Dmatch_survey4}.
Early methods approach this from an axiomatic perspective, incorporating theories such as optimal transport~\cite{opt_trans1,opt_trans2,morphflow} and functional maps~\cite{functionmap1,functionmap2,functionmap3,functionmap4} to compute correspondences within the same category.
In the deep learning era, one line of work \cite{datamap1,datamap2,densematcher} adopts a data-driven paradigm, enabling the prediction of 3D correspondences using curated 3D datasets.
Another category~\cite{featuremap1,featuremap3,featuremap4,featuremap5,gaussianmorphing} leverages powerful 2D feature extractors~\cite{clip,dinov2,stable_diffusion} to establish 3D correspondences in a zero-shot manner.
However, these methods struggle to reliably find correspondences across diverse object categories, limiting the practicality of matching-based 3D morphing.

3DMorpher~\cite{3Dmorpher} improves cross-category morphing performance by incorporating 3DGS-based 3D generative priors. 
However, it cannot handle complex geometries due to limitations in its base generator. 
Moreover, 3DGS-based outputs are not compatible with most commercial 3D software~\cite{blender,maya,unrealengine}. 
In this work, inspired by Trellis~\cite{trellis}—a milestone in 3D generation—we propose a series of simple yet effective modules based on its SLAT representation to achieve aesthetically pleasing and semantically coherent 3D morphing without any training.

\noindent
\textbf{3D Generative Models.}
Advances in 2D diffusion models~\cite{diffusion_ddim,diffusion_ddpm,diffusion_sde,dit,sit,rectified_flow} and the availability of large-scale 3D datasets~\cite{objaverse,objaverse_xl} have driven rapid progress in 3D generation~\cite{lrm,syncdreamer,wonder3d,crm,instantmesh,lgm,unique3d,gaussiancube,3DShape2VecSet,clay,michelangelo,triposr,triposg,triposf,sparc3d,ultra3d,direct3d,direct3d_s2,3dtopia_Xl,hunyuan3d_v1,hunyuan3d_v2,hunyuan3d_v2.5,hunyuan3d_omni,dora,craftsman3d,gaussiananything,trellis,hi3dgen}
Among these, Trellis~\cite{trellis} stands out as a groundbreaking method for native 3D generation.
Its proposed Structured LATent (SLAT) representation not only efficiently encodes rich visual features but is also easily modeled and understood within modern generative frameworks.
Crucially, SLAT possesses a regular and explicit structure, offering superior extensibility compared to implicit or irregular representations such as VecSet~\cite{3DShape2VecSet}, NeRF~\cite{nerf}, or 3DGS~\cite{3dgs}.
A growing body of follow-up work demonstrates that SLAT can be directly transferred—without retraining—to various downstream tasks, including 3D editing~\cite{voxhammer,nano3d}, stylization~\cite{stylesculptor}, correspondence estimation~\cite{HNSR}, articulated object modeling~\cite{FreeArt3D}, scene generation~\cite{hiscene}, and world synthesis~\cite{syncity,TRELLISWorld}.
Despite its strong generalization ability, SLAT has not yet been explored for 3D morphing.
In this paper, we bridge this gap by analyzing fusion rules of SLAT features in cross-/self-attention blocks, and propose lightweight yet solid modules and strategies that fully unlock the potential of SLAT for high-quality, training-free 3D morphing.

\begin{figure*}[tp]
    \centering
    \includegraphics[width=0.95\linewidth]{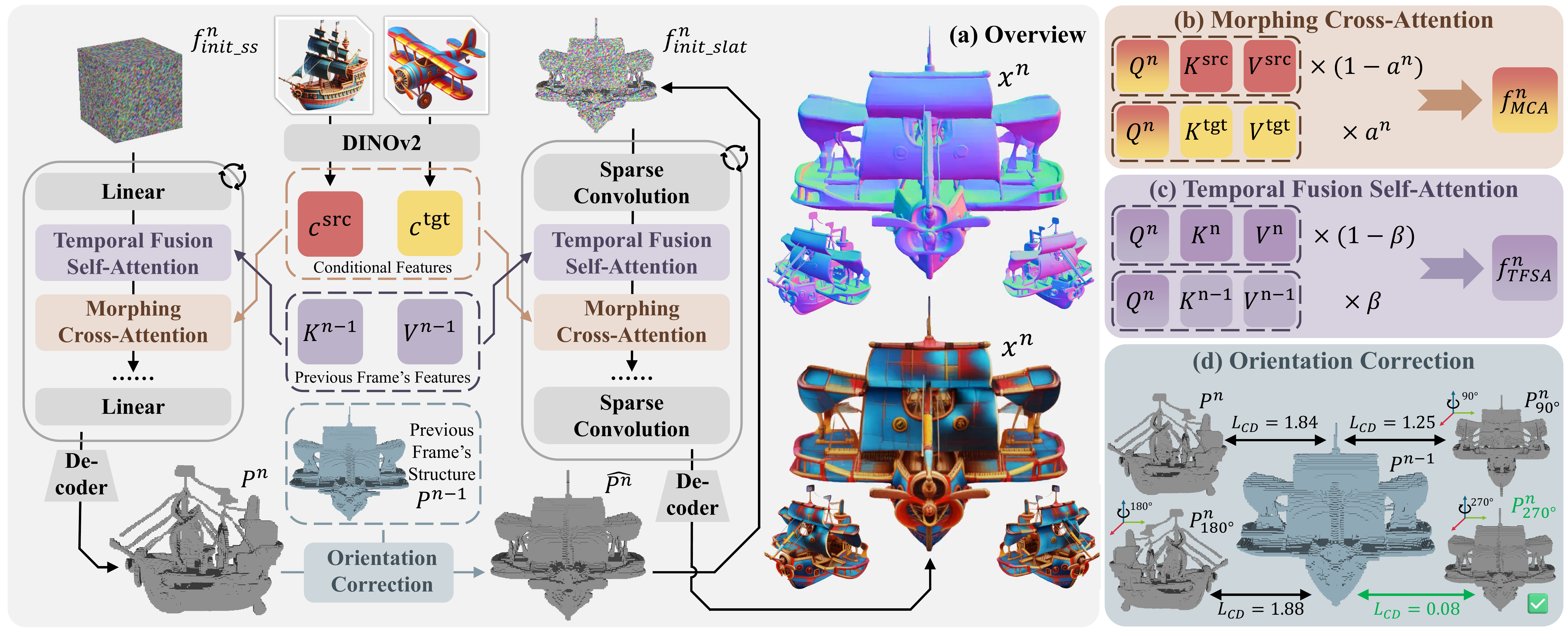} 
    \caption{(a) Overview of our method. MorphAny3D generates a smooth and high-quality morphing sequence between diverse object categories by leveraging the SLAT representation without any training. (b) Morphing Cross-Attention (MCA) fuses information from the source and target objects in the cross-attention layers to ensure the structural coherence and aesthetics of the deformation. (c) Temporal-Fused Self-Attention (TFSA) enhances temporal smoothness by incorporating SLAT features from the previous morphing frame into the self-attention mechanism, enabling smooth transitions over time. (d) An orientation correction strategy inspired by statistical orientation distribution patterns in Trellis-generated assets is proposed to resolve abrupt orientation shifts.}
    \label{fig:pipeline}
    \vspace{-0.6cm}
\end{figure*}

\section{Methodology}
\label{sec:3}

\subsection{Preliminaries}
\label{sec:3.1}
\textbf{Trellis}~\cite{trellis} is a robust feed-forward 3D generative model that produces diverse, high-fidelity 3D assets from text or image prompts.
Its core encodes 3D models into the Structured LATent (SLAT) representation, i.e., a set of local latent vectors $\{(z_i, p_i)\}_{i=1}^L$ anchored at active sparse voxels $p_i$ on the object surface, where each $z_i \in \mathbb{R}^C$ captures fine-scale geometry and appearance.
Trellis employs a two-stage generation pipeline based on rectified flow models~\cite{rectified_flow} tailored for SLAT generation: 
(1) Sparse Structure (SS) Stage: SS flow transformer estimates a $64^{3}$ voxel grid to identify the sparse structure $P=\{p_i\}_{i=1}^L$ representing the global shape of assets.
(2) Structured Latent (SLAT) Stage: SLAT flow transformer predicts the local latent vectors $Z=\{z_i\}_{i=1}^L$ for the voxels identified in the SS stage, producing rich geometry and texture details.
Finally, the generated SLAT is decoded into standard 3D representations such as meshes, NeRF~\cite{nerf}, or 3DGS~\cite{3dgs}.
In this work, we adopt the image-to-3D variant of Trellis~\cite{trellis}, as image conditions facilitate modeling finer details.
Notably, as shown in \cref{sec:4.5}, our method generalizes seamlessly to other SLAT-based models with different input modalities.

\noindent
\textbf{Attention}~\cite{transformer} mechanism is a fundamental component in modern generative models~\cite{dit,sit} and is formulated as:
\begin{equation}
    \text{Attn}(Q, K, V) = \text{Softmax}\left(\frac{QK^T}{\sqrt{d_k}}\right)V,
\end{equation}
where $Q$ is the query features derived from the latent feature $f$, while $K$ and $V$ are the key and value features obtained from either external conditions (cross-attention) or the latent feature itself (self-attention).

\subsection{Overview}
\label{sec:3.2}
Given a source object $x^\text{src}$ and a target object $x^\text{tgt}$, our goal is to generate a smooth, high-quality morphing sequence $\{x^n\}_{n=0}^N$ that gradually transforms $x^\text{src}$ into $x^\text{tgt}$ using the SLAT representation from Trellis~\cite{trellis}.
Both objects may be real 3D assets or Trellis-synthesized.
For real assets, we obtain their initial noised latents $f^\text{src}_\text{init}$ and $f^\text{tgt}_\text{init}$, and image conditions $c^\text{src}$, $c^\text{tgt}$ via 3D inversion~\cite{voxhammer}.
Here, the latent $f=(f_{ss}, f_{slat})$ for each object corresponds to Trellis's SS and SLAT stages.
For Trellis-generated assets, we reuse cached initial features and conditions from the original generation.
The initial noisy feature for frame $n$, $f^n_\text{init}$, is computed by spherical interpolation~\cite{diffmorpher} with a deformation weight $\alpha^n \in [0,1]$, ensuring $x^0 = x^\text{src} (\alpha^0 = 0)$ and $x^N = x^\text{tgt} (\alpha^N = 1)$.
We set $N=49$ (50 frames) with linearly spaced $\alpha^n = n / N$.
Further details are provided in our Supp. Mat.
\cref{fig:pipeline}-(a) shows the MorphAny3D overview.
Based on observed SLAT feature blending patterns in cross-/self-attention blocks (\cref{sec:3.3}, \cref{fig:comp_fusion}), we introduce two key components to improve morphing plausibility and temporal coherence: Morphing Cross-Attention (\cref{sec:3.4}, \cref{fig:pipeline}-(b)) and Temporal-Fused Self-Attention  (\cref{sec:3.5}, \cref{fig:pipeline}-(c)).
We also propose an orientation correction strategy (\cref{sec:3.6}, \cref{fig:pipeline}-(d)) derived from statistical analysis of orientation distributions in Trellis outputs to suppress abrupt pose shifts.

\subsection{SLAT Fusion Patterns in Attention Mechanism}
\label{sec:3.3}
\begin{figure}[tp]
    \centering
    \includegraphics[width=0.45\textwidth]{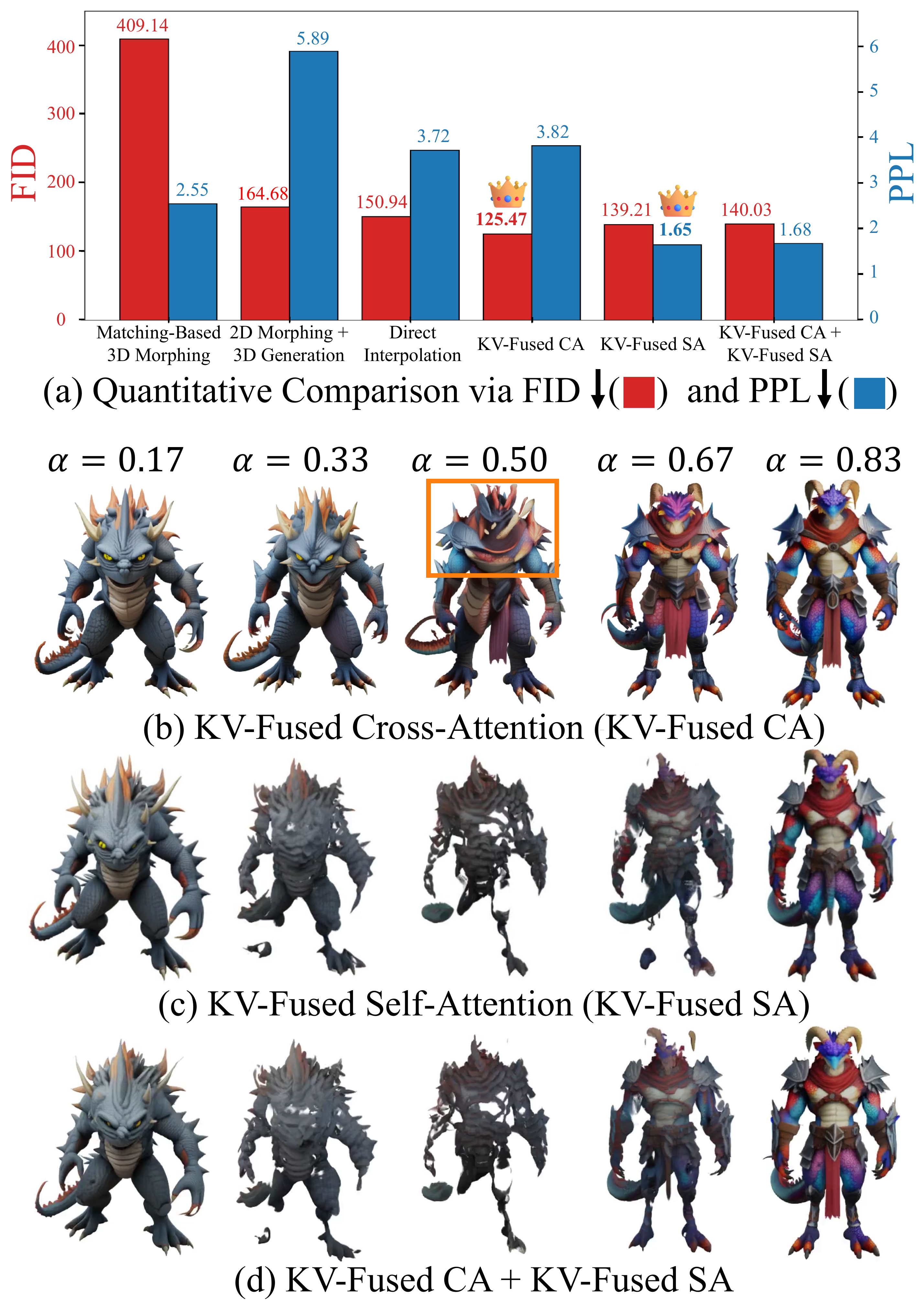} 
    \caption{Analysis of SLAT fusion patterns in attention for 3D morphing. (a) FID (plausibility) and PPL (smoothness) comparison. (b, c, d) Qualitative results of different fusion strategies (same case as \cref{fig:comp}).}
    \label{fig:comp_fusion}
    \vspace{-0.6cm}
\end{figure}

As shown in \cref{fig:comp}, naively applying SLAT to 3D morphing yields poor transitions, highlighting the need to understand how SLAT features influence the morphing process.
Prior work~\cite{diffmorpher,3Dmorpher} shows that fusing source and target keys and values in attention greatly improves 2D/3D morphing quality and continuity:
\begin{equation}
    \begin{split}
    &\text{KV-Fused-Attn}(Q^{n}, K^{\text{src/tgt}}, V^{\text{src/tgt}}) = \text{Attn}\big(Q^{n}, (1- \\ 
    &\alpha^{n}) K^{\text{src}} + \alpha^{n} K^{\text{tgt}}, (1-\alpha^{n}) V^{\text{src}} + \alpha^{n} V^{\text{tgt}} \big),
    \end{split}
\end{equation}
where $Q^{n}$ comes from frame-$n$ latent features $f^{n}$, and $K^{\text{src/tgt}}$ and $V^{\text{src/tgt}}$ are derived from image conditions (cross-attention) or latent features (self-attention) of source and target objects, respectively.
This fusion is shared across all diffusion timesteps and generation stages; we omit corresponding indices for brevity.
Given the poor performance of naive SLAT interpolation and the success of KV fusion, we extend this strategy to Trellis's cross-attention (CA) and self-attention (SA), yielding KV-Fused CA and KV-Fused SA.
We quantitatively evaluate methods on a test set using FID (lower = more plausible) and PPL (lower = smoother) (\cref{fig:comp_fusion}-(a); see \cref{sec:4.1} for evaluation details) and qualitatively compare fusion strategies in \cref{fig:comp_fusion}-(b, c, d).

From these results, we draw three key observations:
(1) KV-Fused CA significantly enhances the structural and semantic plausibility of 3D morphing by fusing 2D conditions from the source and target objects within cross-attention (the lowest FID). 
However, it fails to eliminate localized implausible artifacts, as shown in the orange box of \cref{fig:comp_fusion}-(b).
(2) KV-Fused SA effectively improves the smoothness and continuity of the morphing sequence by aggregating 3D latent features from the source and target objects within self-attention (the lowest PPL).
(3) Yet, simultaneously applying KV-fused CA and KV-fused SA—while aiming to maximize both plausibility and smoothness—compromises the plausibility of the resulting deformation (see \cref{fig:comp_fusion}-(d)).
These results indicate that while SLAT feature blending benefits morphing, naive combination harms the plausibility-smoothness trade-off.
To address this, we propose two tailored modules—Morphing Cross-Attention and Temporal-Fused Self-Attention—that preserve the advantages of fusion while avoiding its pitfalls.

\subsection{Morphing Cross-Attention (MCA)}
\label{sec:3.4}
\begin{figure}[tp]
    \centering
    \includegraphics[width=0.45\textwidth]{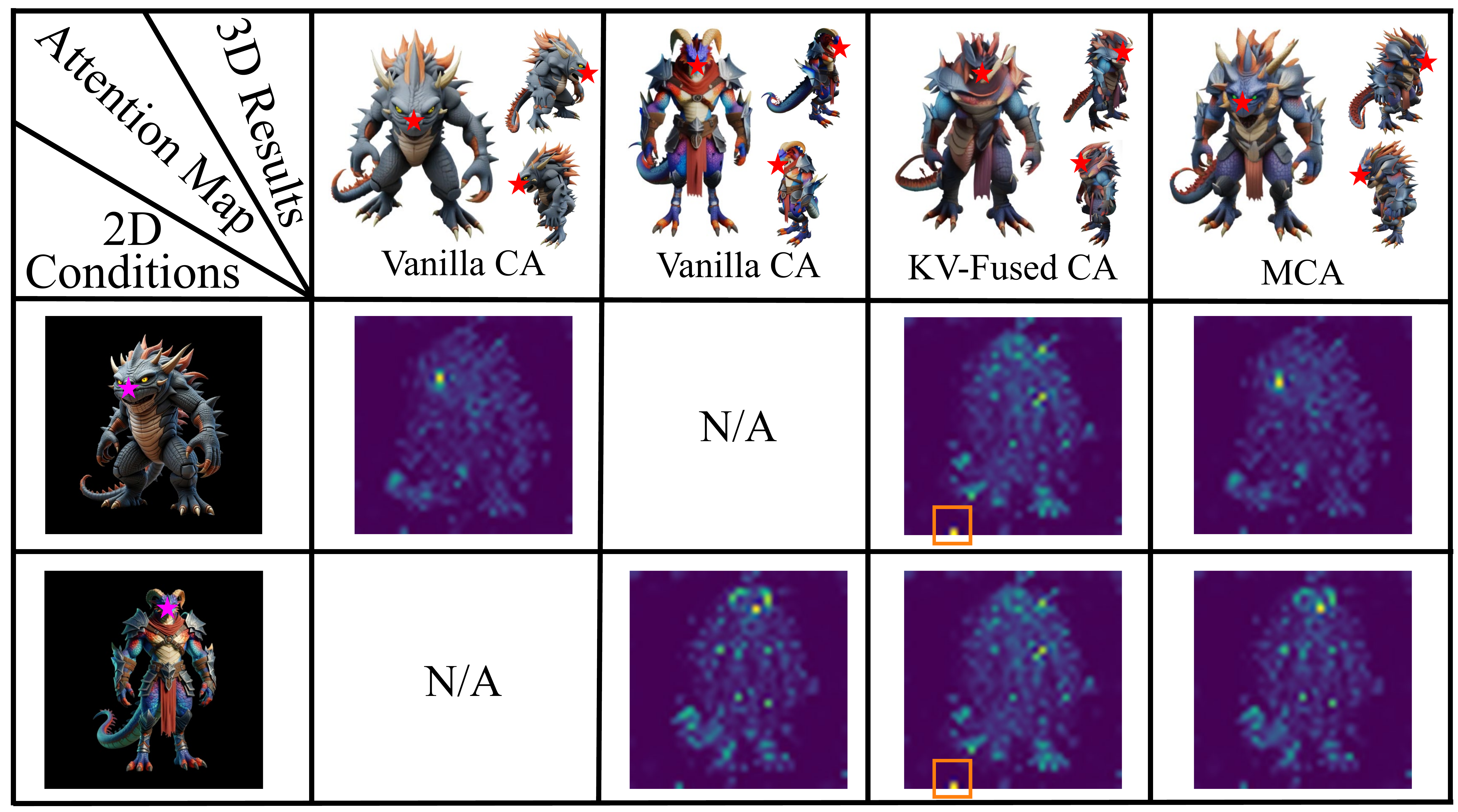} 
    \caption{Attention maps visualization for different attention mechanisms. Red stars denote head SLAT features; pink stars mark their corresponding input regions. Orange boxes highlight KV-Fused CA's incorrect attention focus. MCA preserves correct, semantically consistent attention and avoids KV-Fused CA's artifacts shown in \cref{fig:comp_fusion}-(b).}
    \label{fig:attention_map}
    \vspace{-0.45cm}
\end{figure}

We hypothesize that the local irrational structure in KV-Fused CA is caused by patch-wise blending of source and target 2D conditions, which introduces ambiguous semantics that mislead the generation model.
Specifically, the blended keys and values in cross-attention are derived from patch-wise DINOv2~\cite{dinov2} features.
However, spatially aligned patches between source and target images often lack semantic correspondence, leading to distorted outputs.
To verify this hypothesis, we visualize the cross-attention maps of head SLAT features (marked by red stars) in \cref{fig:attention_map}.
The attention maps from vanilla CA (second and third columns) correctly focus on the head regions in the input conditions (indicated by pink stars). 
In contrast, KV-Fused CA (fourth column) erroneously attends to background regions (highlighted by an orange box), subsequently using semantically mismatched features to guide head SLAT generation—resulting in local structural distortions.
To address this issue, we propose Morphing Cross-Attention (MCA) (\cref{fig:pipeline}-(b)).
Instead of blending keys and values prior to the attention computation, MCA computes separate attention outputs for the source and target objects and then combines them to produce the final feature $f^{n}_{MCA}$:
\begin{equation}
    \begin{split}
    &\text{MCA}(Q^{n}, K^{\text{src/tgt}}, V^{\text{src/tgt}}) = (1-\alpha^{n}) \text{Attn}\big(Q^{n}, \\
    &K^{\text{src}}, V^{\text{src}}\big) + \alpha^{n} \text{Attn}\big(Q^{n}, K^{\text{tgt}}, V^{\text{tgt}}\big).
    \end{split}
\end{equation}
As shown in the last column of \cref{fig:attention_map}, MCA preserves accurate attention on semantically consistent regions by processing source and target features independently, thereby avoiding the artifacts observed in KV-Fused CA.
This improvement is further corroborated in \cref{sec:4.3}.

\begin{figure}[tp]
    \centering
    \includegraphics[width=0.45\textwidth]{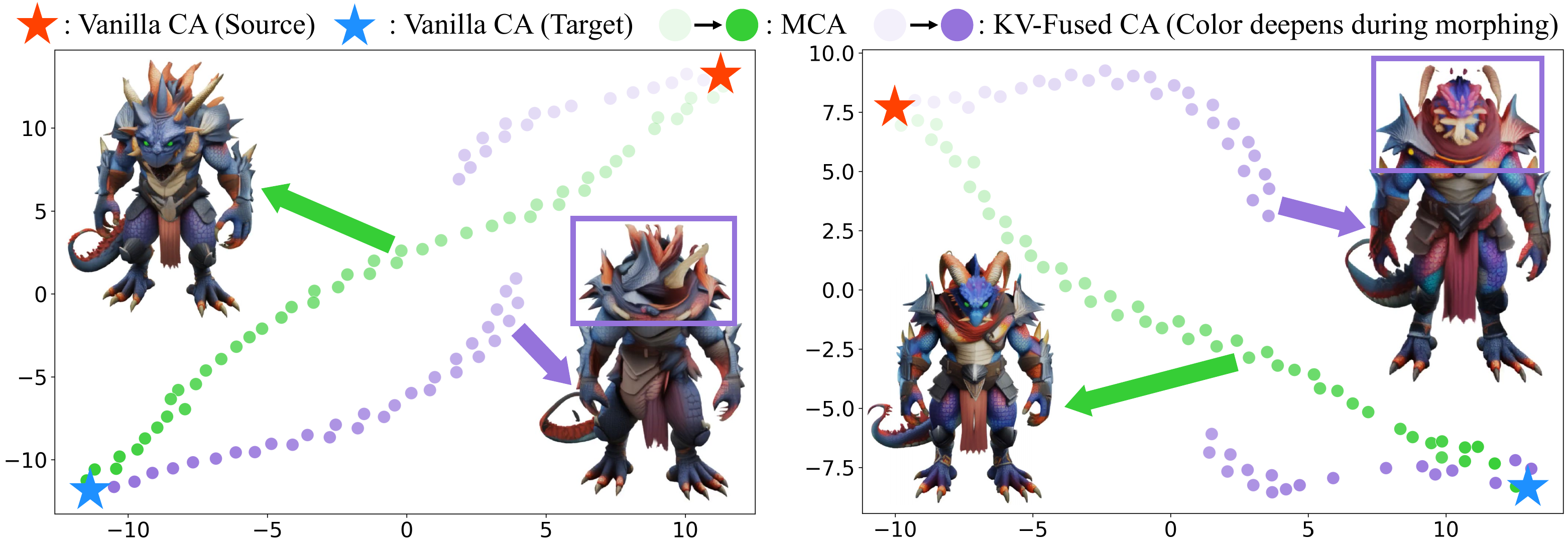} 
    \caption{t-SNE visualization of attention features for different attention mechanisms in the SS Stage (left) and SLAT Stage (right). Obviously, MCA exhibits a stable, smooth feature trajectory, whereas KV-Fused CA's is erratic and interrupted.}
    \label{fig:t-sne}
    \vspace{-0.6cm}
\end{figure}

Furthermore, we verified MCA's efficacy via feature distribution analysis.
\cref{fig:t-sne} shows t-SNE~\cite{t-sne} visualizations of average output features from vanilla CA, MCA, and KV-Fused CA across all blocks and diffusion timesteps in the SS and SLAT stages (green/purple deepen during morphing). 
Notably, MCA exhibits a stable, smooth trajectory, whereas KV-Fused CA yields an erratic, interrupted path.
Combined with \cref{fig:attention_map}, this reveals that naive KV fusion disrupts conditions' semantic consistency, causing unstable 3D morphing (interrupted purple trajectory) and chaotic outputs (purple arrow). 
In contrast, MCA preserves accurate, semantically consistent attention maps, enabling smooth, plausible 3D deformations (stable green trajectory and arrow).

\subsection{Temporal-Fused Self-Attention (TFSA)}
\label{sec:3.5}
Although MCA yields structurally coherent and visually plausible deformations, temporal smoothness remains limited due to the frame-wise independence inherent in standard 3D generators.
To address this limitation—and building on insights from KV-Fused SA discussed in \cref{sec:3.3}—we propose Temporal-Fused Self-Attention (TFSA).
As illustrated in \cref{fig:pipeline}-(c), TFSA promotes temporally consistent 3D morphing by incorporating features from previously generated frames into the self-attention mechanism. 
Specifically, when generating the $n$ frame, TFSA fuses attention outputs from current and previous frames' keys and values to produce the output $f^{n}_{TFSA}$:
\begin{equation}
    \begin{split}
    &\text{TFSA}(Q^{n}, K^{n}, V^{n}, K^{n-1}, V^{n-1}) = (1-\beta) \text{Attn}\big(\\
    &Q^{n}, K^{n}, V^{n}\big) + \beta \text{Attn}\big(Q^{n}, K^{n-1}, V^{n-1}),
    \end{split}
\end{equation}
where $\beta \in [0, 1]$ controls the influence of the prior frame.
We empirically set $\beta=0.2$.
Unlike the KV-Fused SA in \cref{fig:comp_fusion}-(e, f)—which blends source and target features—TFSA fuses features from already plausible neighboring morphing frames, preserving semantic fidelity while enhancing smoothness, as shown in \cref{sec:4.3} and \cref{fig:ablation}-(b).

\begin{figure}[tp]
    \centering
    \includegraphics[width=0.95\linewidth]{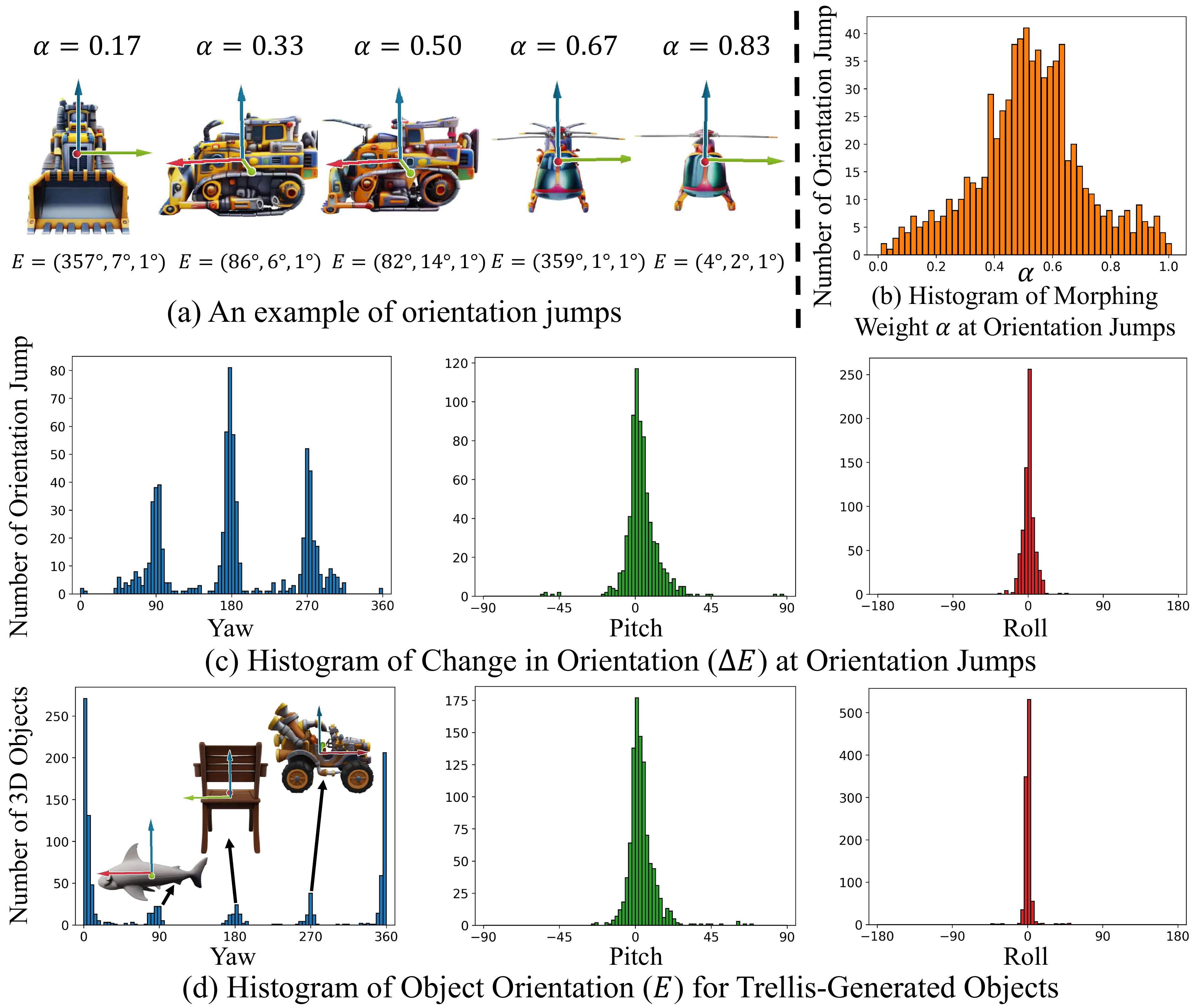}
    \caption{(a) Example of abrupt orientation change during morphing. (b) Distribution of $\alpha$ at orientation jumps, peaking near intermediate stages. (c) Adjacent-frame orientation changes $\Delta E$ at orientation jumps, dominated by 90°, 180°, and 270° yaw shifts. (d) Orientation distribution of Trellis-generated objects, showing non-canonical poses clustered at the same yaw angles.}
    \label{fig:orientation_correction}
    \vspace{-0.6cm}
\end{figure}

\subsection{Orientation Correction}
\label{sec:3.6}
Morphing often suffers from abrupt orientation changes (i.e., orientation jumps)—especially in intermediate stages—causing visually jarring transitions (\cref{fig:orientation_correction}-(a)).
We represent orientation via the Z-Y-X Euler angles $E=(\text{yaw}, \text{pitch}, \text{roll})$ (in degrees), estimated using OrientAnything~\cite{OrientAnything}.
An orientation jump is flagged when the adjacent-frame angular difference $\Delta E$ exceeds 45°, a threshold accounting for estimation noise.

Analyzing 200 sequences (50 frames each) generated with MCA and TFSA, we find that jumps concentrate around $\alpha\approx0.5$ (\cref{fig:orientation_correction}-(b)), where the conditional features are the most ambiguous.
Moreover, $\Delta E$ is strongly biased toward yaw rotations of 90°, 180°, and 270°, with negligible pitch/roll changes (\cref{fig:orientation_correction}-(c)), suggesting a systematic bias rather than randomness.
We hypothesize that this stems from Trellis's internal pose prior.
To verify, we estimate the orientations of 1,000 Trellis-generated objects (\cref{fig:orientation_correction}-(d)): while most adopt canonical poses, non-canonical ones cluster precisely at the same yaw angles (see cases in \cref{fig:orientation_correction}-(d)-Yaw)—confirming the link between orientation jumps and Trellis's learned pose distribution.

Based on these observations, we propose a lightweight orientation correction strategy to mitigate abrupt orientation changes (\cref{fig:pipeline}-(d)).
After generating the sparse structure $P^{n}$ in the SS stage, we create four yaw-rotated candidates: $P^{n}$, $P^{n}_{90^{\circ}}$, $P^{n}_{180^{\circ}}$, and $P^{n}_{270^{\circ}}$.
We select the candidate that minimizes the Chamfer Distance (CD) to the previous frame's structure $P^{n-1}$ as the corrected structure $\hat{P^{n}}$, which is then passed to the SLAT flow transformer.
Since jumps occur mainly mid-sequence (\cref{fig:orientation_correction}-(b)), this leverages early stable poses to correct later errors. 
When no jump occurs, the unrotated $P^{n}$ yields the lowest CD and is retained—ensuring non-intrusiveness. 
As shown in \cref{sec:4.3}, this significantly improves temporal smoothness without sacrificing fidelity.

\begin{figure*}[tp]
    \centering
    \includegraphics[width=0.9\linewidth]{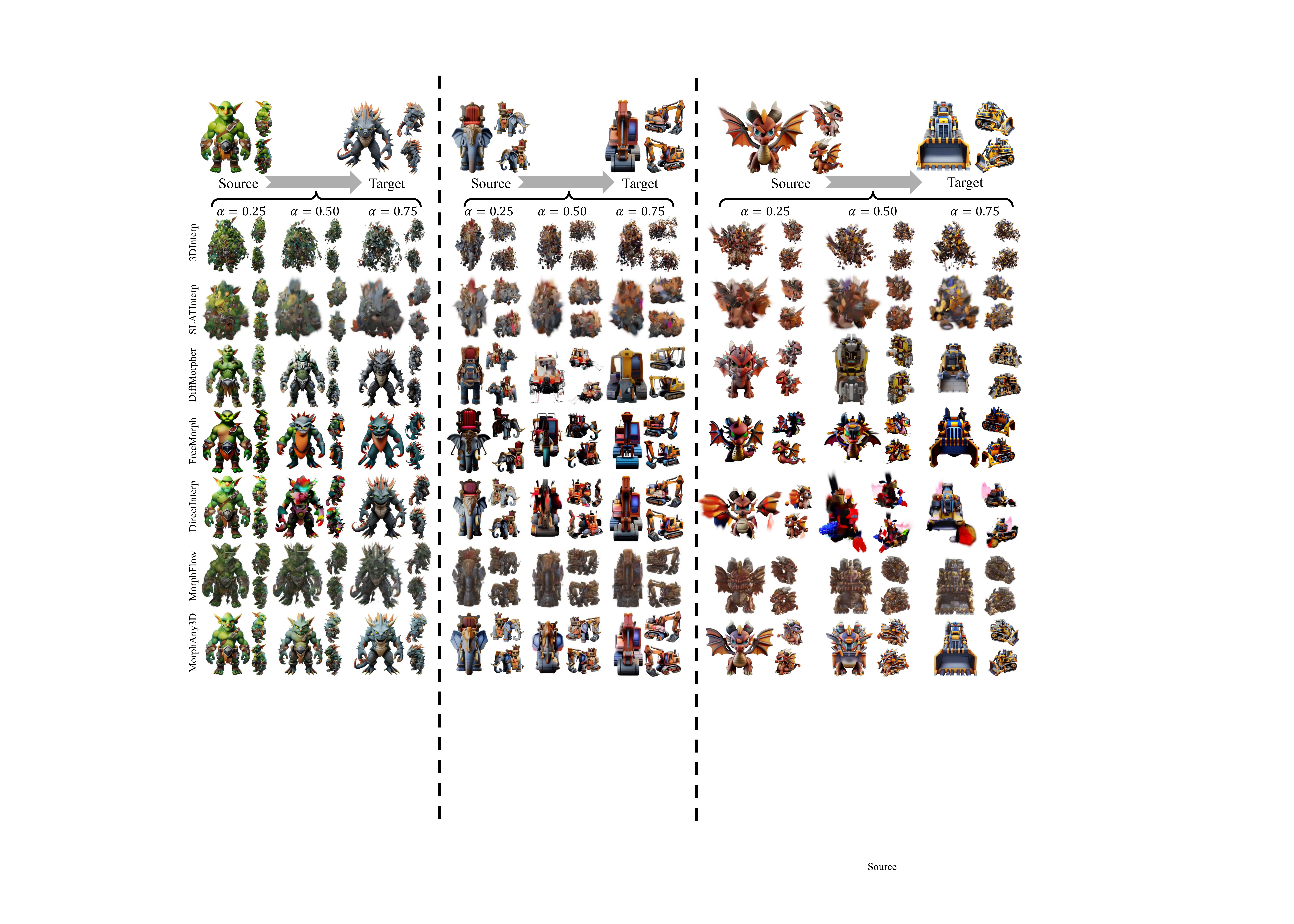}
    \caption{Qualitative comparisons. MorphAny3D generates smooth, high-quality 3D morphing sequences across diverse object categories. \textbf{More results are provided in our Supp. Mat.}}
    \label{fig:qual}
    \vspace{-0.6cm}
\end{figure*}

\section{Experiments}
\label{sec:4}
Due to space limits, we report key settings and results; \textbf{full details are in the Supp. Mat.}
We render RGB views from 3DGS and normal maps from Mesh representations.

\subsection{Experimental Settings}
\label{sec:4.1}
\textbf{Implementation Details.}
Our method uses Image-to-3D Trellis~\cite{trellis} without retraining or hyperparameter tuning. Experiments were conducted on a single A6000 GPU.
Generating each frame takes 30s with 24GB of memory.

\noindent\textbf{Baselines.}
We compare four types:
(1) Matching-based 3D morphing of 3D or SLAT features (\textit{3DInterp}, \textit{SLATInterp}, and use DenseMatcher~\cite{densematcher} for correspondence);
(2) 2D morphing (DiffMorpher~\cite{diffmorpher}, FreeMorph~\cite{freemorph}) is lifted to 3D via Trellis;
(3) Direct interpolation of noise and conditions (\textit{DirectInterp});
(4) Modern 3D morphing method—MorphFlow~\cite{morphflow}. 
Since 3DMorpher~\cite{3Dmorpher} has not released its full code, we provide only a qualitative comparison in our Supp. Mat.

\noindent\textbf{Evaluation Metrics.}
We evaluate 50 diverse source-target pairs from real 3D datasets~\cite{voxhammer,gso} and Trellis-generated assets using four metrics:
(a) FID~\cite{fid} measures visual plausibility by comparing rendered frames to originals;
(b) Perceptual Path Length (PPL) and Perceptual Distance Variance (PDV)~\cite{diffmorpher} assess transition smoothness and temporal homogeneity via average perceptual change and its variance between adjacent frames;
(c) Aesthetics Scores (AS) uses vision-language models~\cite{chatgpt,gemini} to rank visual appeal;
(d) User Preference (UP) is gathered through a user study evaluating quality, smoothness, and realism.

\subsection{Evaluation}
\label{sec:4.2}
\textbf{Qualitative Results.}
As shown in \cref{fig:qual}, MorphAny3D generates smooth, high-quality 3D morphing sequences across diverse categories, outperforming all baselines.
Matching-based methods (e.g., 3DInterp, SLATInterp, MorphFlow) yield smooth, but often implausible or semantically incoherent transitions.
2D-first approaches (DiffMorpher, FreeMorph) leverage strong 2D priors for structural plausibility but suffer from poor 3D temporal consistency, leading to jerky sequences.
DirectInterp's feature fusion is ill-suited to SLAT, resulting in suboptimal deformations.
In contrast, our method unleashes the potential of 3D generative priors in SLAT to produce temporally smooth, semantically meaningful, and visually faithful morphs.
For example, in the elephant-to-excavator case (\cref{fig:qual}), it implicitly aligns the trunk with the boom, creating a coherent hybrid that seamlessly blends both concepts.

\begin{table}[tp]
    \caption{Quantitative comparison. Best and second-best in \textbf{bold} and \underline{underlined}.} 
    \label{tab:quantitative_comparisons}
    \centering
    \renewcommand{\arraystretch}{0.8}
    \resizebox{0.95\linewidth}{!}{
        \begin{tabular}{cccccc}
            \toprule
            Method & FID $\downarrow$ & PPL $\downarrow$ & PDV $\downarrow$ & AS (\%) $\uparrow$ & UP (\%) $\uparrow$ \\
            \midrule
            3DInterp~\cite{densematcher} & 409.14 & 2.55 & \textbf{0.0006} & 1.00 & 0.61 \\
            SLATInterp~\cite{densematcher} & 348.31 & 6.53 & 0.0010 & 0.00 & 1.43 \\
            DiffMorpher~\cite{diffmorpher} & 208.08 & 6.65 & 0.0021 & 5.00 & 0.82 \\
            FreeMorph~\cite{freemorph} & 164.68 & 5.89 & 0.0027 & \underline{11.00} & 3.27 \\
            DirectInterp & \underline{150.94} & 3.72 & 0.0039 & 2.00 & \underline{5.51} \\
            MorphFlow~\cite{morphflow} & 284.96 & \textbf{2.41} & \underline{0.0009} & 0.00 & 1.63 \\
            MorphAny3D & \textbf{111.95} & \underline{2.47} & \textbf{0.0006} & \textbf{81.00} & \textbf{86.73} \\
            \bottomrule
        \end{tabular}
    }
    \vspace{-0.3cm}
\end{table}

\noindent\textbf{Quantitative Results.}
As shown in \cref{tab:quantitative_comparisons}, MorphAny3D achieves state-of-the-art performance, obtaining the best scores in FID, PDV, AS, and UP, and the second-best PPL.
Matching-based methods achieve smoothness via linear interpolation but suffer from poor plausibility, as evidenced by an extremely high FID.
Our method significantly improves plausibility and visual quality with only a marginal smoothness trade-off—its PPL is just 0.06 above the lowest.

\begin{figure}[tp]
    \centering
    \includegraphics[width=0.95\linewidth]{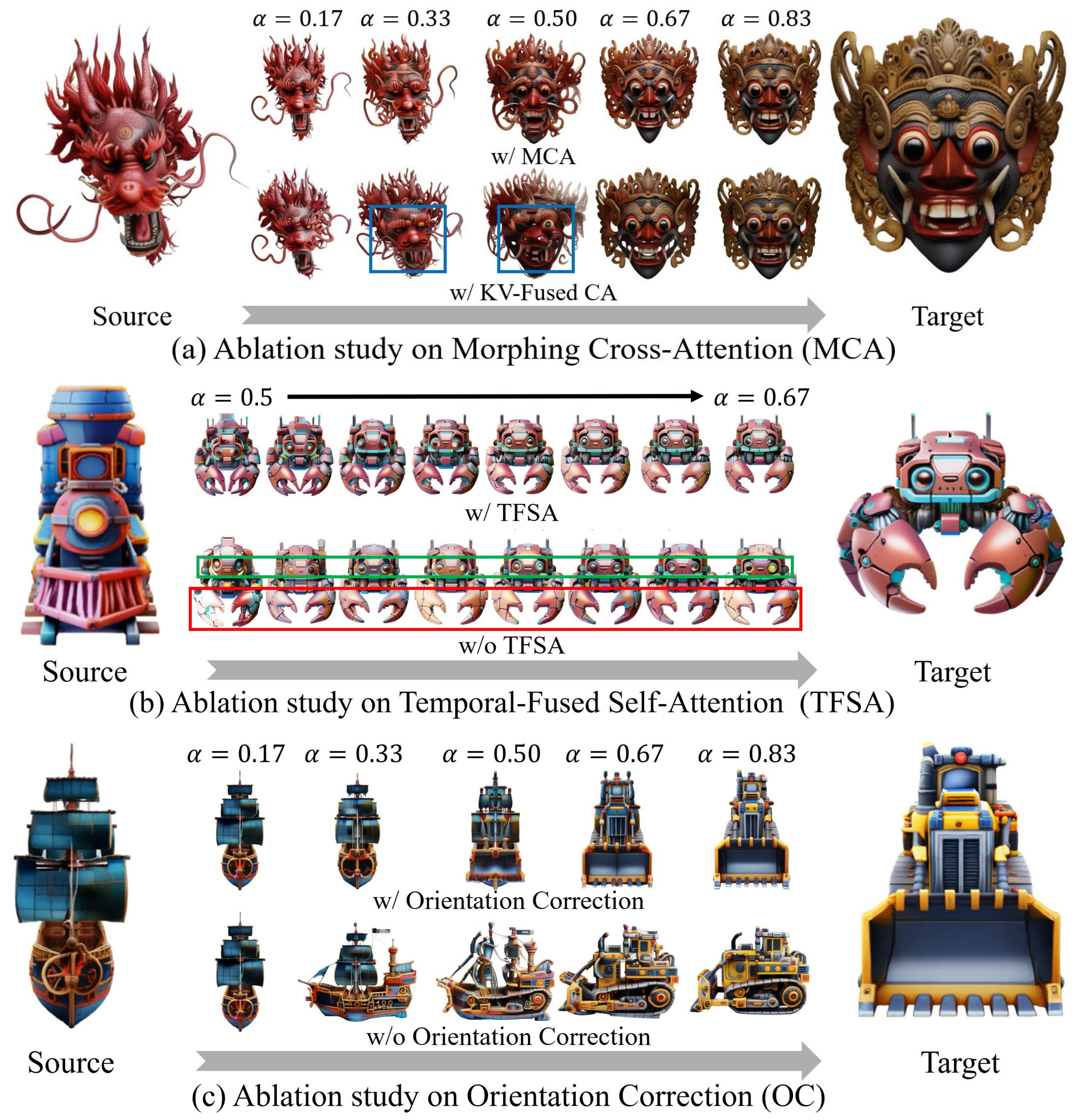}
    \caption{Ablation study on (a) MCA, (b) TFSA, and (c) OC.}
    \label{fig:ablation}
    \vspace{-0.3cm}
\end{figure}

\begin{table}[tp]
    \caption{Ablation study on key components of MorphAny3D.}
    \label{tab:ablation_study}
    \centering
    \renewcommand{\arraystretch}{0.8}
    \resizebox{0.7\linewidth}{!}{
        \begin{tabular}{cccc}
            \toprule
            Method & FID $\downarrow$ & PPL $\downarrow$ & PDV $\downarrow$\\
            \midrule
            KV-Fused CA & 125.47 & 3.82 & 0.0013 \\
            MCA & 112.18 & 3.66 & 0.0010 \\
            MCA + TFSA & \underline{113.22} & \underline{2.87} & \underline{0.0007} \\
            MCA + TFSA + OC & \textbf{111.95} & \textbf{2.47} & \textbf{0.0006} \\
            \bottomrule
        \end{tabular}
    }
    \vspace{-0.6cm}
\end{table}

\subsection{Ablation Study}
\label{sec:4.3}
\textbf{Effectiveness of MCA.}
\cref{fig:ablation}-(a) shows that MCA suppresses local artifacts (blue box) in KV-Fused CA and improves plausibility.
FID drops from 125.47 to 112.18 (\cref{tab:ablation_study}).
\textbf{Effectiveness of TFSA.}
\cref{fig:ablation}-(b) shows improved temporal coherence—e.g., stable crab claws and eyes (green/red boxes). PPL and PDV decrease to 2.87 and 0.0007.
\textbf{Effectiveness of Orientation Correction (OC).}
\cref{fig:ablation}-(c) shows that OC mitigates orientation jumps. Quantitatively, it further reduces PPL to 2.47 and PDV to 0.0006.

\subsection{Applications}
\label{sec:4.4}
\begin{figure}[tp]
    \centering
    \includegraphics[width=0.95\linewidth]{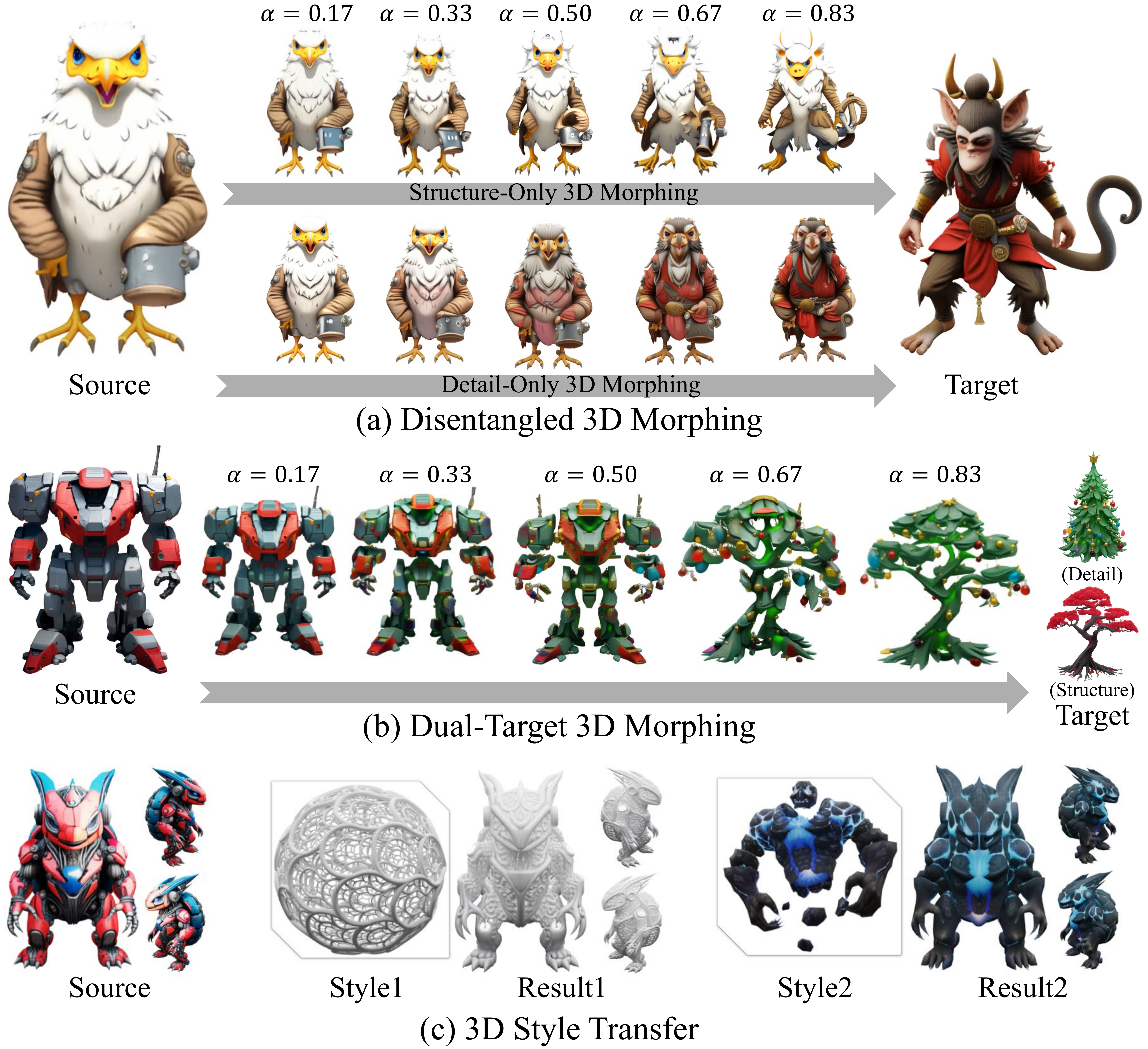}
    \caption{Applications: (a) Disentangled 3D morphing, (b) Dual-Target 3D Morphing and (c) 3D Style Transfer.}
    \label{fig:app}
    \vspace{-0.3cm}
\end{figure}

\textbf{Disentangled 3D Morphing.}
By applying our method selectively to Trellis's SS and SLAT stages, we decouple the global structure from local details. 
\cref{fig:app}-(a) shows sequences that preserve the source structure with target details (or vice versa).
\textbf{Dual-Target 3D Morphing.}
Assigning different targets to SS and SLAT enables morphing toward two concepts simultaneously.
\cref{fig:app}-(b) shows one target shaping structure, the other detail.
\textbf{3D Style Transfer.}
Using a style image as the SLAT target transfers style while preserving source structure and aesthetics (\cref{fig:app}-(c)).

\subsection{Generalization Ability}
\label{sec:4.5}
\begin{figure}[tp]
    \centering
    \includegraphics[width=0.95\linewidth]{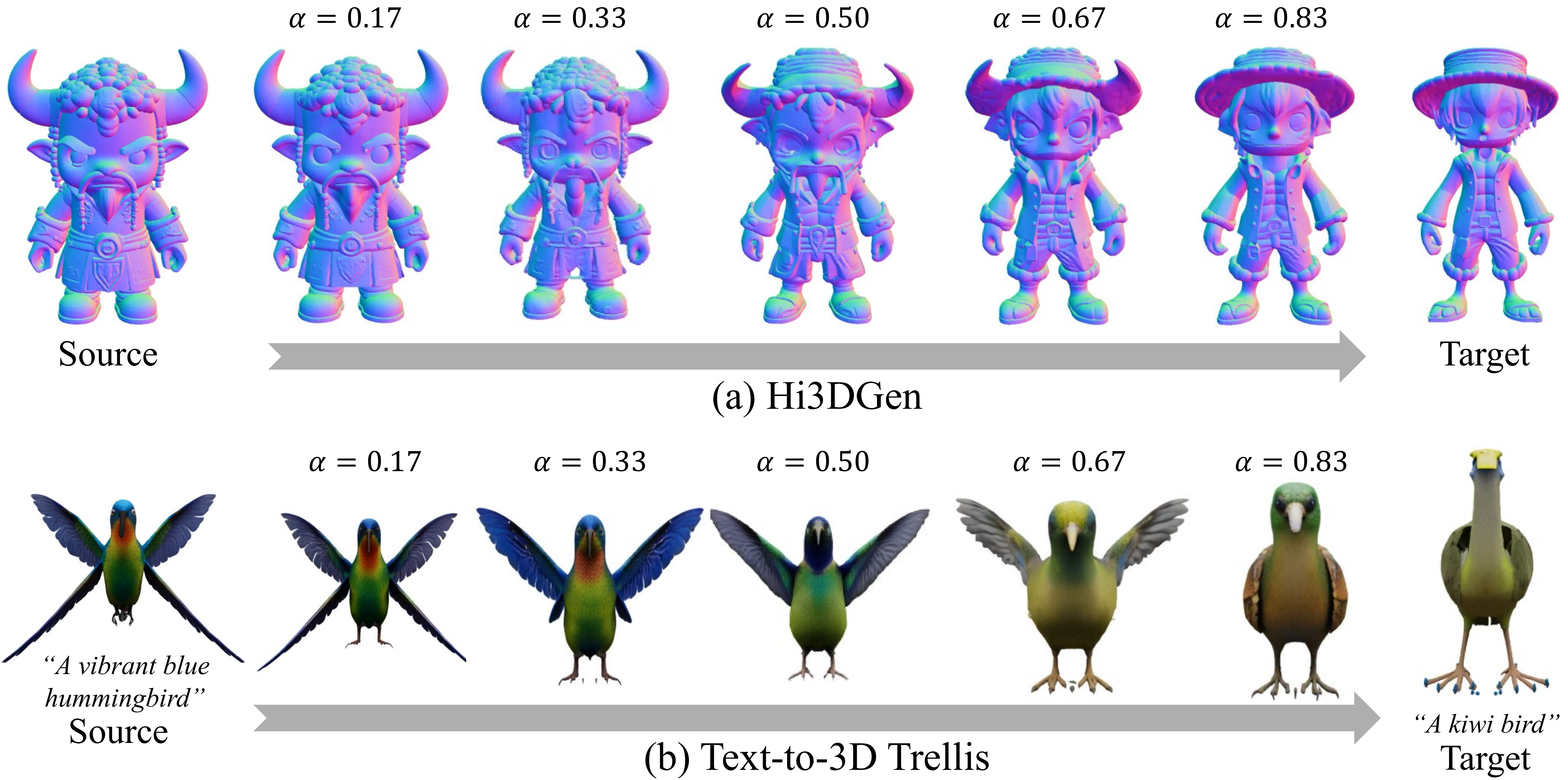}
    \caption{Generalization experiments on (a) Hi3DGen and (b) Text-to-3D Trellis.}
    \label{fig:general}
    \vspace{-0.3cm}
\end{figure}

We apply MorphAny3D to other SLAT-based models—Hi3DGen~\cite{hi3dgen} and Text-to-3D Trellis~\cite{trellis}. 
As shown in \cref{fig:general}, it consistently produces smooth, high-quality morphs, confirming its versatility.

\begin{figure}[tp]
    \centering
    \includegraphics[width=0.95\linewidth]{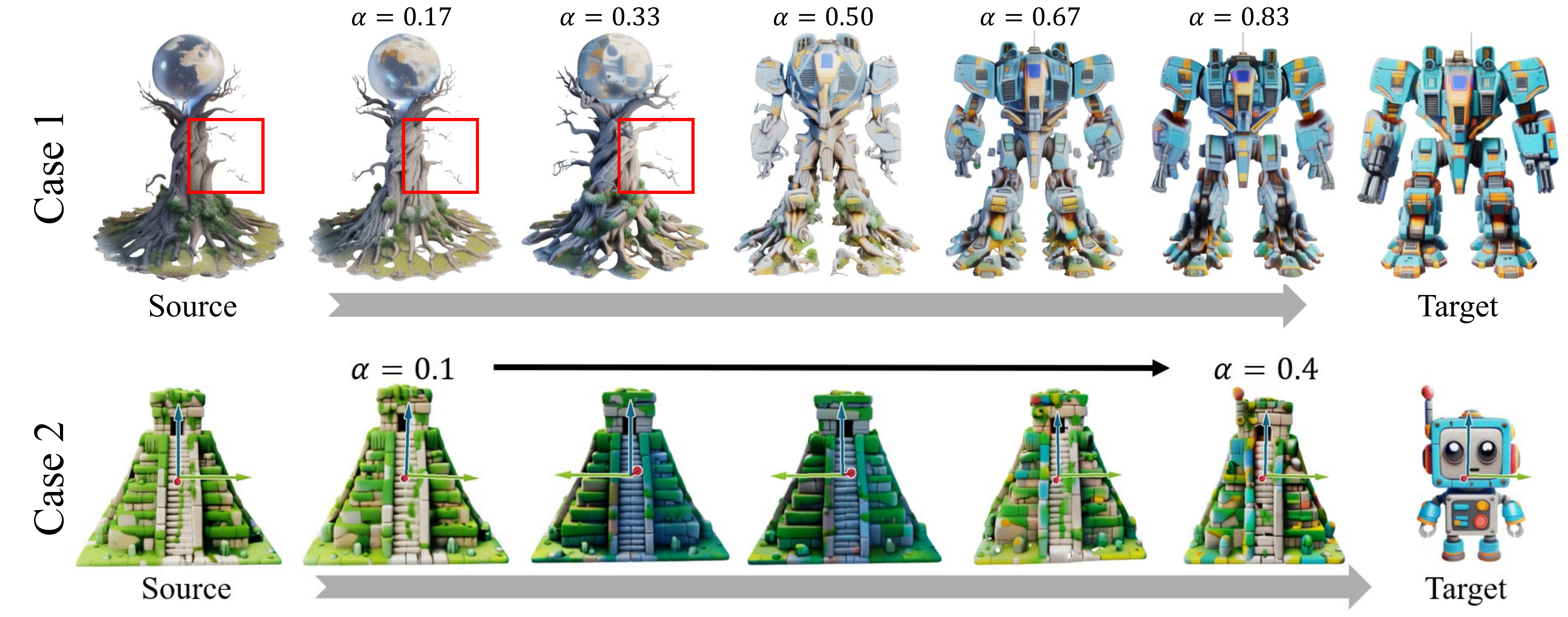}
    \caption{Failure cases.}
    \label{fig:failcase}
    \vspace{-0.7cm}
\end{figure}

\section{Conclusion}
\label{sec:5}
We present MorphAny3D, a training-free 3D morphing framework based on Trellis's SLAT representation.
By analyzing SLAT fusion in attention, we introduce Morphing Cross-Attention (MCA)—fusing source and target features for structural and semantic fidelity—and Temporal-Fused Self-Attention (TFSA)—leveraging prior frames for temporal coherence.
We also propose an orientation correction strategy, guided by statistical pose patterns in Trellis outputs, to reduce abrupt orientation shifts.
Experiments show that our method achieves state-of-the-art results in generating smooth, high-quality 3D morphing sequences across diverse object categories.
It further supports advanced applications such as decoupled morphing and 3D style transfer, and readily generalizes to other SLAT-based models.

\noindent\textbf{Limitations and Future Work.}
MorphAny3D inherits Trellis's limitations and may exhibit artifacts on extremely fine structures (see red boxes of case 1 in \cref{fig:failcase}).
Future work could adopt stronger 3D generative backbones~\cite{triposf} or enhance SLAT representations to capture finer geometric details.
Additionally, our orientation correction strategy may miss yaw-symmetric objects' rotations (see case 2 in \cref{fig:failcase}).
Finally, the high runtime restricts its application; we plan to investigate acceleration methods like KV cache.

\noindent\textbf{Acknowledgement.}
This work was supported by the NSFC under Grant No. 62376121, Basic Research Program of Jiangsu under Grant No. BK20251999, Gusu Innovation Leading Talent Program under Grant No. ZXL2025319, and Jiangsu Provincial Science \& Technology Major Project under Grant No. BG2024042.

{
    \small
    \bibliographystyle{ieeenat_fullname}
    \bibliography{main}
}

\clearpage

\section{Supplemental Material Overview}
This document includes the following supplementary sections:
\begin{itemize}
    \item Initialization Details.
    \item Additional Experimental Details.
    \item Extended 3D Morphing Results.
    \item Additional Qualitative Experiments.
    \item Additional Application Details.
    \item Generalization Details.
\end{itemize}

\section{Initialization Details}
\textbf{3D Inversion.}
For real-world assets, initial noisy latents \(f_{\text{init}}\) and image conditions \(c\) are obtained via 3D inversion, following the procedure in VoxHammer~\cite{voxhammer}. 
This involves two stages: Sparse Structure (SS) and Structured Latent (SLAT), which map a 3D asset to its corresponding initial latents \(f_{\text{init\_ss}}\) and \(f_{\text{init\_slat}}\). 
Further implementation details are provided in Sec. 3.2 of VoxHammer~\cite{voxhammer}.

\noindent
\textbf{Initial Features Interpolation.}
The initial feature for frame $n$, $f^n_\text{init}$, is computed by spherical interpolation~\cite{diffmorpher} with a deformation weight $\alpha^n \in [0,1]$, ensuring $x^0 = x^\text{src} (\alpha^0 = 0)$ and $x^N = x^\text{tgt} (\alpha^N = 1)$.
Specifically, we obtain the initial features in the SS stage of source and target assets via 3D inversion, denoted as $f_\text{init\_ss}^\text{src}$ and $f_\text{init\_ss}^\text{tgt}$.
The initial feature in the SS stage for frame $n$ is calculated as:
\begin{equation}
    f^n_\text{init\_ss} = \frac{\sin((1-\alpha^n)\theta)}{\sin(\theta)} f_\text{init\_ss}^\text{src} + \frac{\sin(\alpha^n \theta)}{\sin(\theta)} f_\text{init\_ss}^\text{tgt},
    \label{eq:slerp}
\end{equation}
where $\theta = \arccos(\frac{f_\text{init\_ss}^\text{src} \cdot f_\text{init\_ss}^\text{tgt}}{\|f_\text{init\_ss}^\text{src}\| \|f_\text{init\_ss}^\text{tgt}\|})$.
A similar approach is used for the SLAT stage, but requires finding corresponding sparse voxels based on Euclidean distance before performing spherical interpolation.

\begin{figure}[tp]
    \centering
    \includegraphics[width=0.9\linewidth]{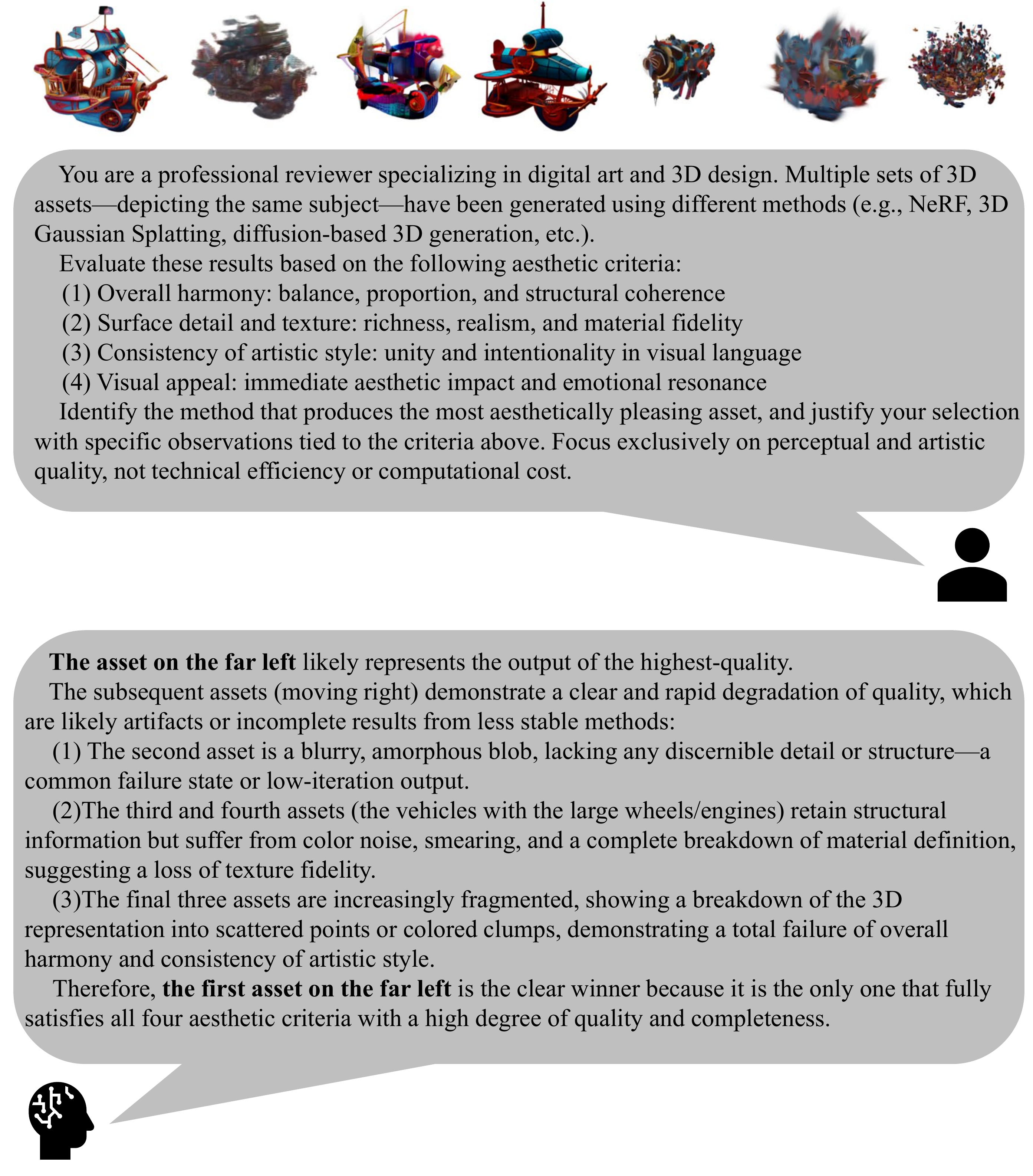} 
    \caption{Examples of Aesthetics Score (AS) evaluation using vision-language models Gemini-2.5~\cite{gemini} and ChatGPT-5~\cite{chatgpt}. Given rendered images of morphed 3D assets, the models select the most aesthetically pleasing result across all methods.}
    \label{fig:vqa}
    \vspace{-0.6cm}
\end{figure}

\section{Additional Experimental Details}
\textbf{Implementation Details.}
Our method builds upon Image-to-3D Trellis~\cite{trellis}, replacing all cross-attention modules with our proposed Morphing Cross-Attention (MCA) modules and self-attention modules with Temporal-Fused Self-Attention (TFSA) modules. 
After obtaining the structure of the n-th frame $P^{n}$, orientation correction is performed. 
All modules introduced require no training, allowing flexible and cost-effective application to other SLAT-based models. 
Both stages use a sampling step of 25 and a classifier-free guidance (CFG) guidance scale of 5.0.

\noindent
\textbf{Baselines Implementation Details.}
(1) 3DInterp and SLTAInterp: We first establish correspondence between the target and source objects using a well-known 3D shape correspondence method, DenseMatcher~\cite{densematcher} (ICLR 2025, Spotlight). We then perform interpolation directly on the corresponding 3D-representation or SLAT-representation to generate the morphed 3D results.
(2) DiffMorpher~\cite{diffmorpher} and FreeMorph~\cite{freemorph}: We use their officially released code to obtain 2D morphing results. Subsequently, we generate the final 3D assets from the morphed images using Image-to-3D Trellis~\cite{trellis}.
(3) MorphFlow~\cite{morphflow}: We first render multi-view images of the source and target 3D assets. Their official code is then used to generate the final morphed results.

\begin{figure*}[tp]
    \centering
    \includegraphics[width=0.98\linewidth]{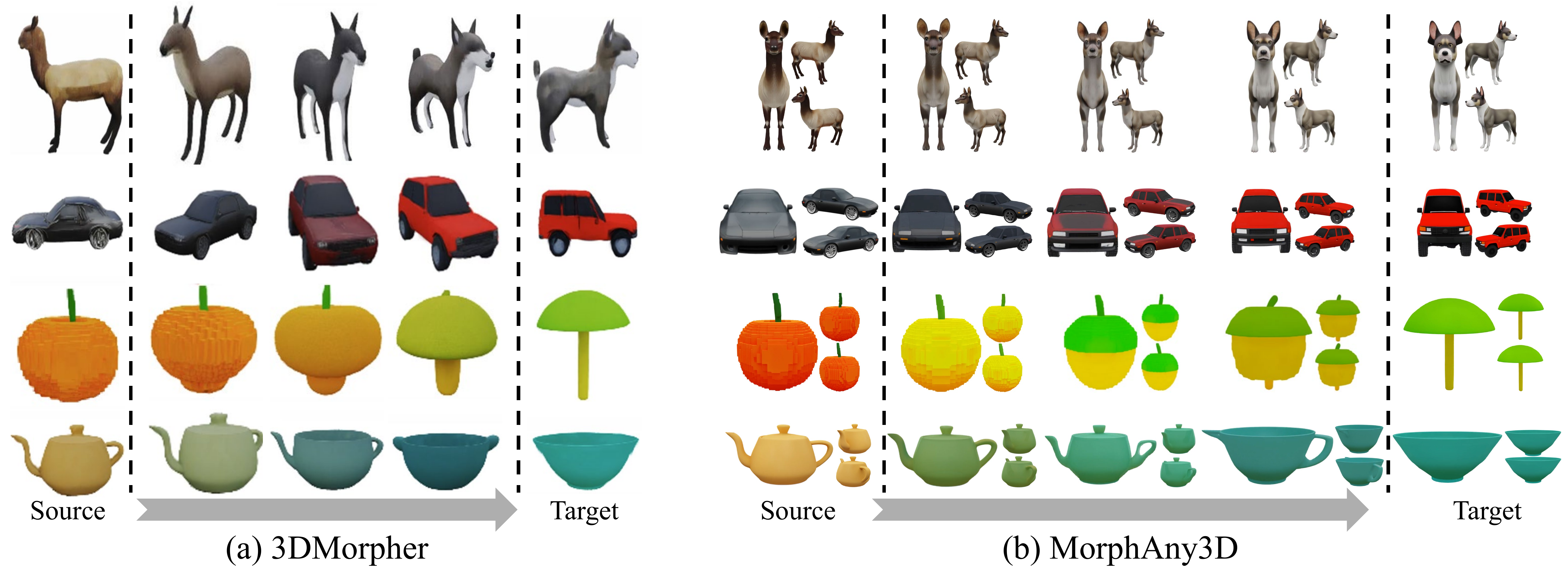} 
    \caption{Qualitative comparison with 3DMorpher~\cite{3Dmorpher}. Results from 3DMorpher are reproduced directly from their published paper.}
    \label{fig:comp3}
    \vspace{-0.3cm}
\end{figure*}

\begin{figure*}[tp]
    \centering
    \includegraphics[width=0.98\linewidth]{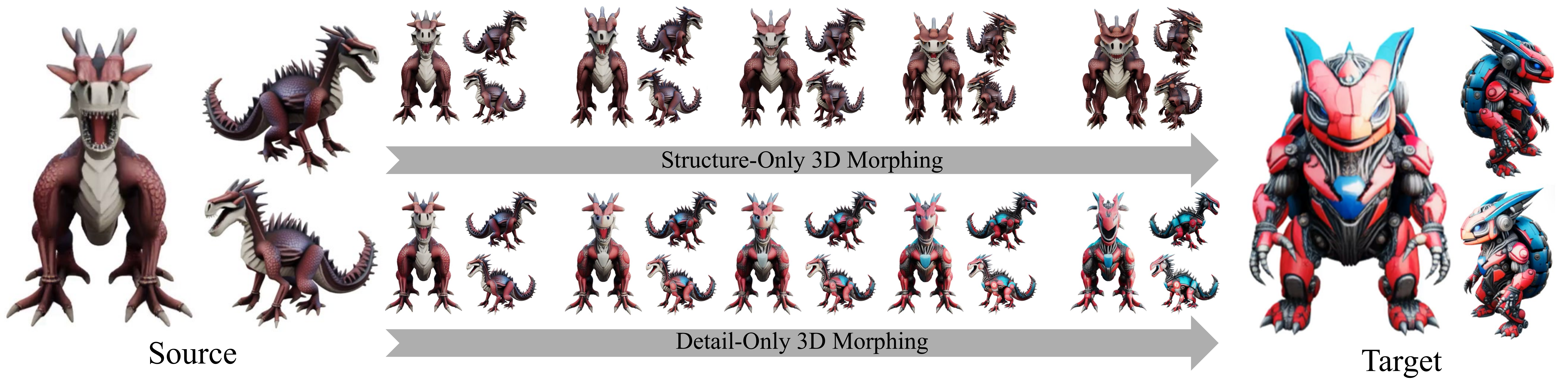} 
    \caption{Disentangled 3D morphing. By applying our Morphing Cross-Attention (MCA) and Temporal-Fused Self-Attention (TFSA) modules exclusively in one stage (SS or SLAT), we independently control structural and detail deformation.}
    \label{fig:app1}
    \vspace{-0.6cm}
\end{figure*}

\begin{figure*}[tp]
    \centering
    \includegraphics[width=0.98\linewidth]{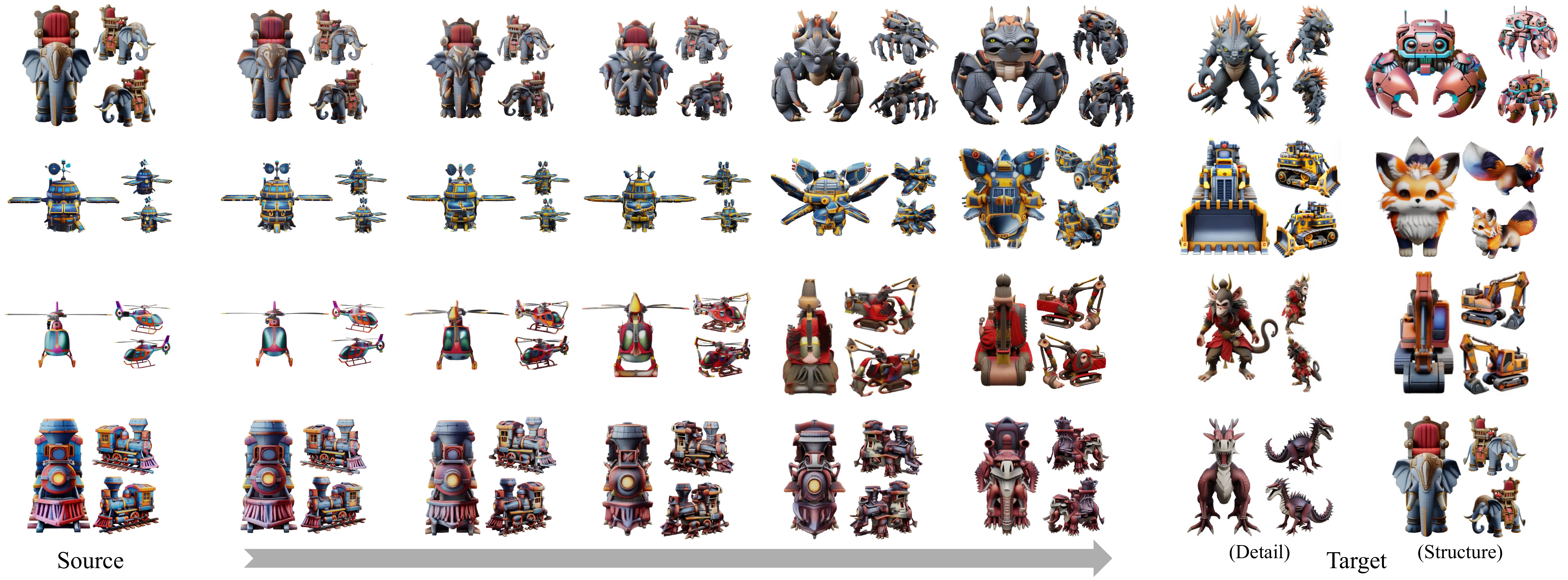} 
    \caption{Dual-target 3D morphing. Distinct target assets are assigned to the SS and SLAT stages, enabling flexible composition of global shape and local details from two different sources.}
    \label{fig:app2}
    \vspace{-0.3cm}
\end{figure*}

\begin{figure*}[tp]
    \centering
    \includegraphics[width=0.98\linewidth]{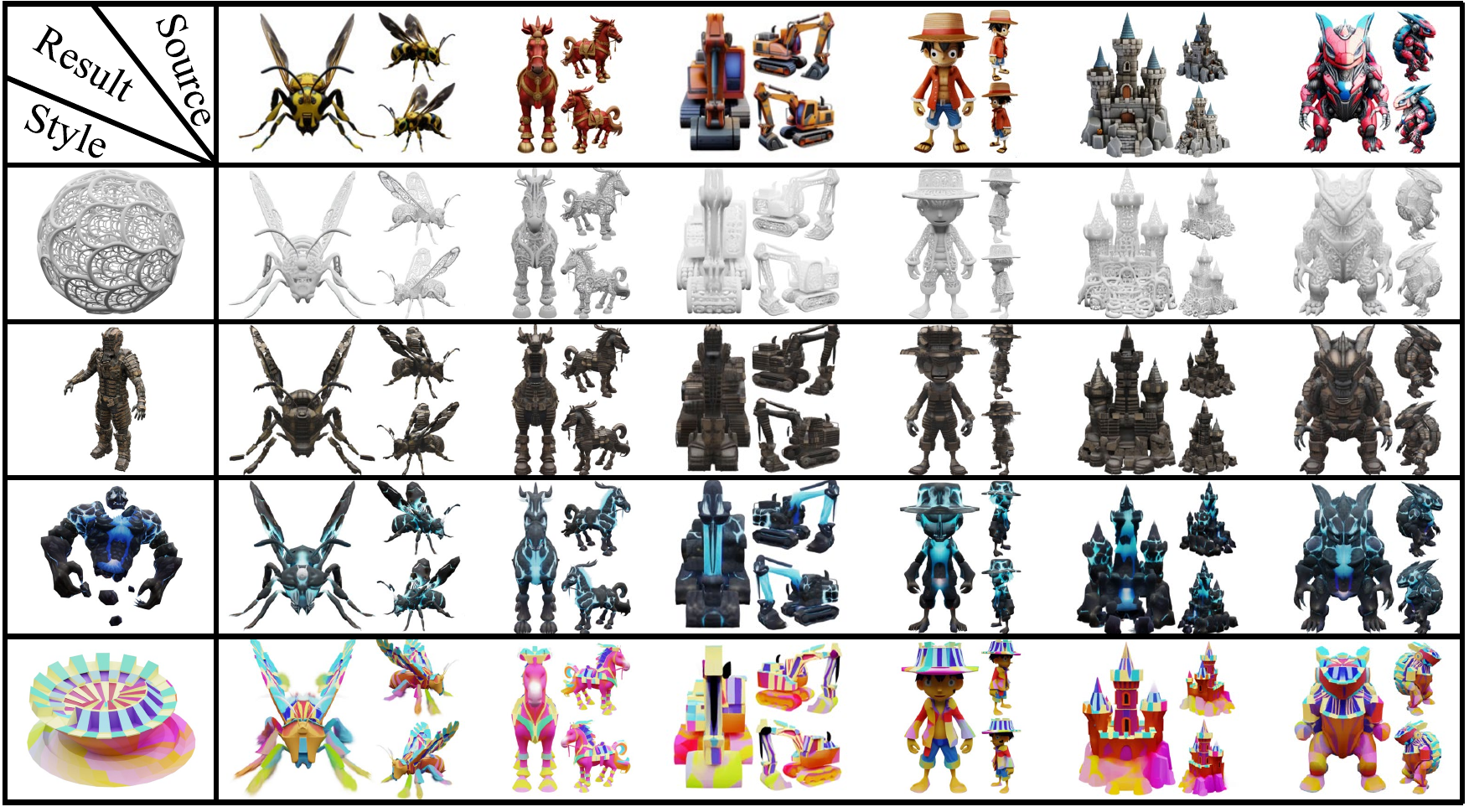} 
    \caption{3D style transfer. A style reference guides the SLAT stage while the SS stage retains the source structure, allowing texture and geometric style to be transferred without altering the underlying shape.}
    \label{fig:app3}
    \vspace{-0.6cm}
\end{figure*}

\begin{figure*}[tp]
    \centering
    \includegraphics[width=0.8\linewidth]{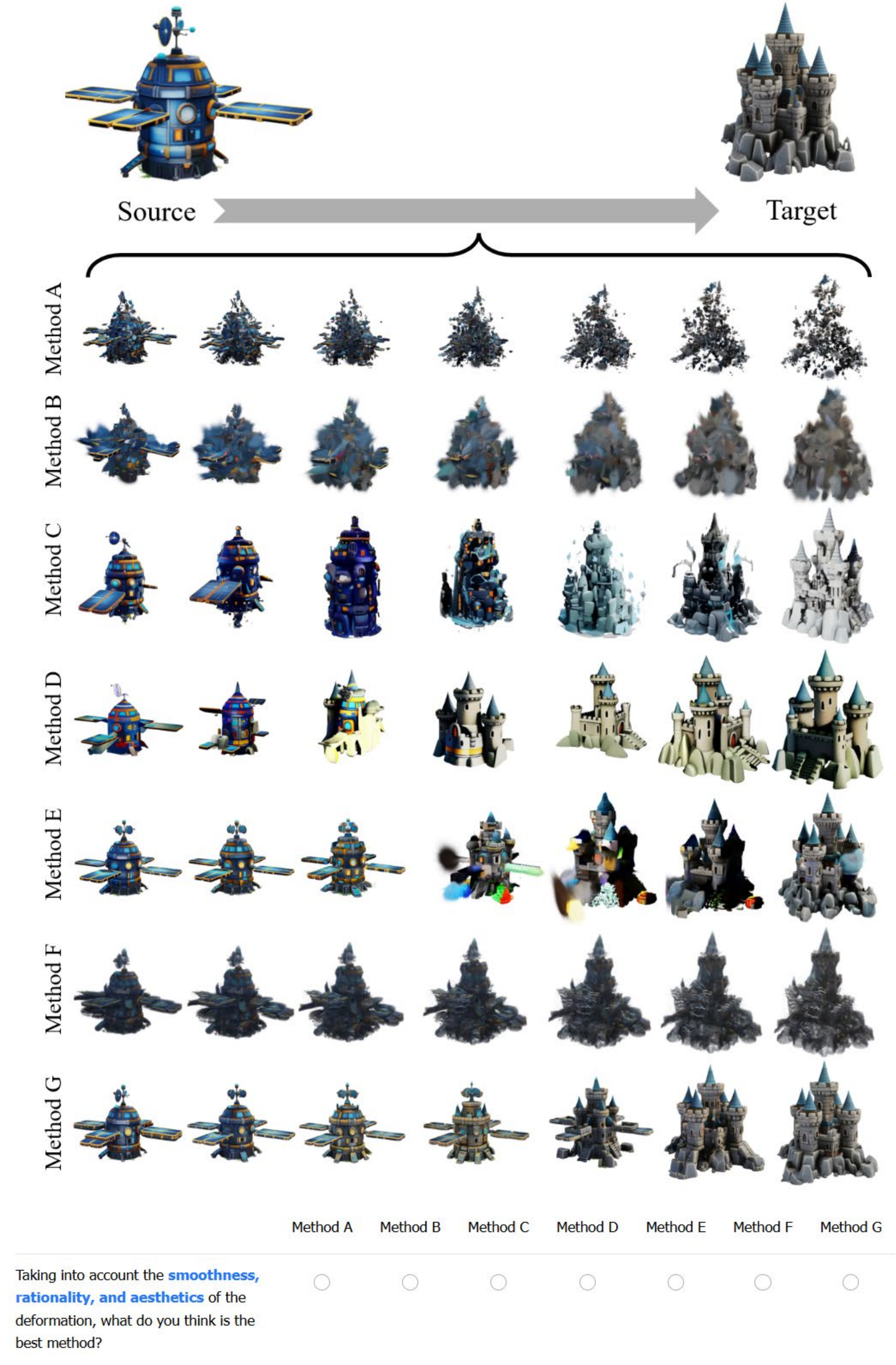} 
    \caption{User Preference (UP) evaluation. Participants selected their preferred morphed 3D asset from all methods based on rendered images, judging quality, smoothness, and realism.}
    \label{fig:user}
    \vspace{-0.6cm}
\end{figure*}

\noindent
\textbf{Evaluation Metrics Details.}
(1) Fréchet Inception Distance (FID)~\cite{fid}: For each morphed asset and its originals, we render 12 images uniformly sampled along the yaw axis and compute FID between the rendered distributions.
(2) Perceptual Path Length (PPL) and Perceptual Distance Variance (PDV)~\cite{diffmorpher}: For each source-target pair, we render six morphing videos (uniform yaw views) and compute PPL/PDV as in DiffMorpher~\cite{diffmorpher}.
(3) Aesthetics Scores (AS): Using Gemini-2.5~\cite{gemini} and ChatGPT-5~\cite{chatgpt}, we prompt the models to select the most visually appealing morphed asset per pair (see \cref{fig:vqa}). AS is the selection frequency (as a percentage).
(4) User Preference (UP): We conducted a user study with 49 participants experienced in computer vision and graphics. 
They selected their preferred result across 10 source-target pairs based on quality, smoothness, and realism (see \cref{fig:user}). UP is reported as the selection percentage.

\section{Extended 3D Morphing Results}
Additional results are shown in \cref{fig:show1,fig:show2}. 
Our method produces high-quality, smooth morphs across diverse categories—including animals, vehicles, buildings, and humanoids. 
We visualize RGB renders from the 3DGS representation and normal maps from the mesh to highlight both appearance and geometric fidelity.

\section{Additional Qualitative Experiments}
\begin{figure*}[tp]
    \centering
    \includegraphics[width=0.85\linewidth]{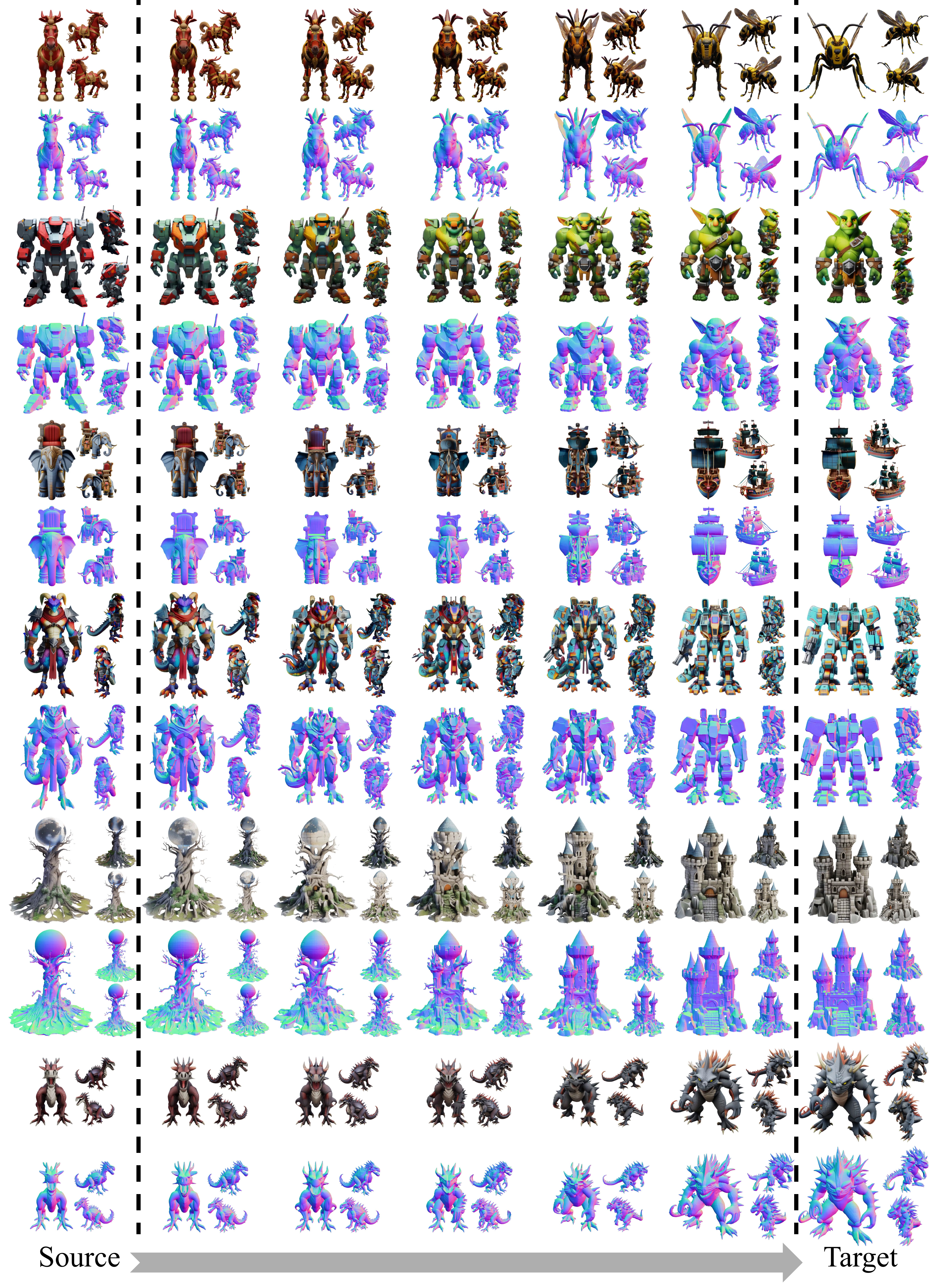} 
    \caption{Extended 3D morphing results across diverse object categories (e.g., animals, vehicles, buildings). Our method produces high-fidelity and temporally smooth transitions.}
    \label{fig:show1}
    \vspace{-0.6cm}
\end{figure*}

\begin{figure*}[tp]
    \centering
    \includegraphics[width=0.78\linewidth]{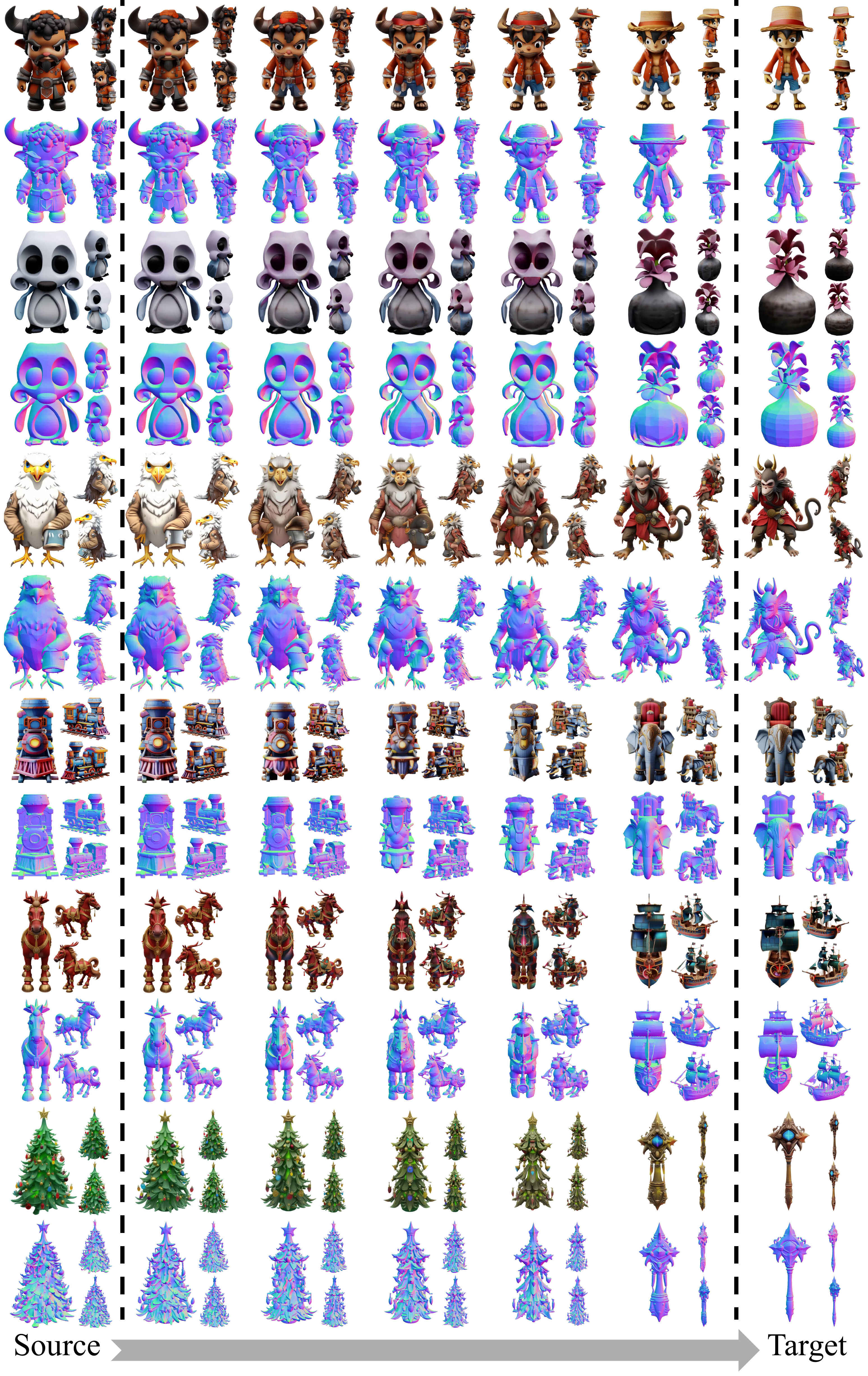} 
    \caption{Additional extended results demonstrating consistent morphing quality across varied and challenging object categories.}
    \label{fig:show2}
    \vspace{-0.6cm}
\end{figure*}

\begin{figure*}[tp]
    \centering
    \includegraphics[width=0.9\linewidth]{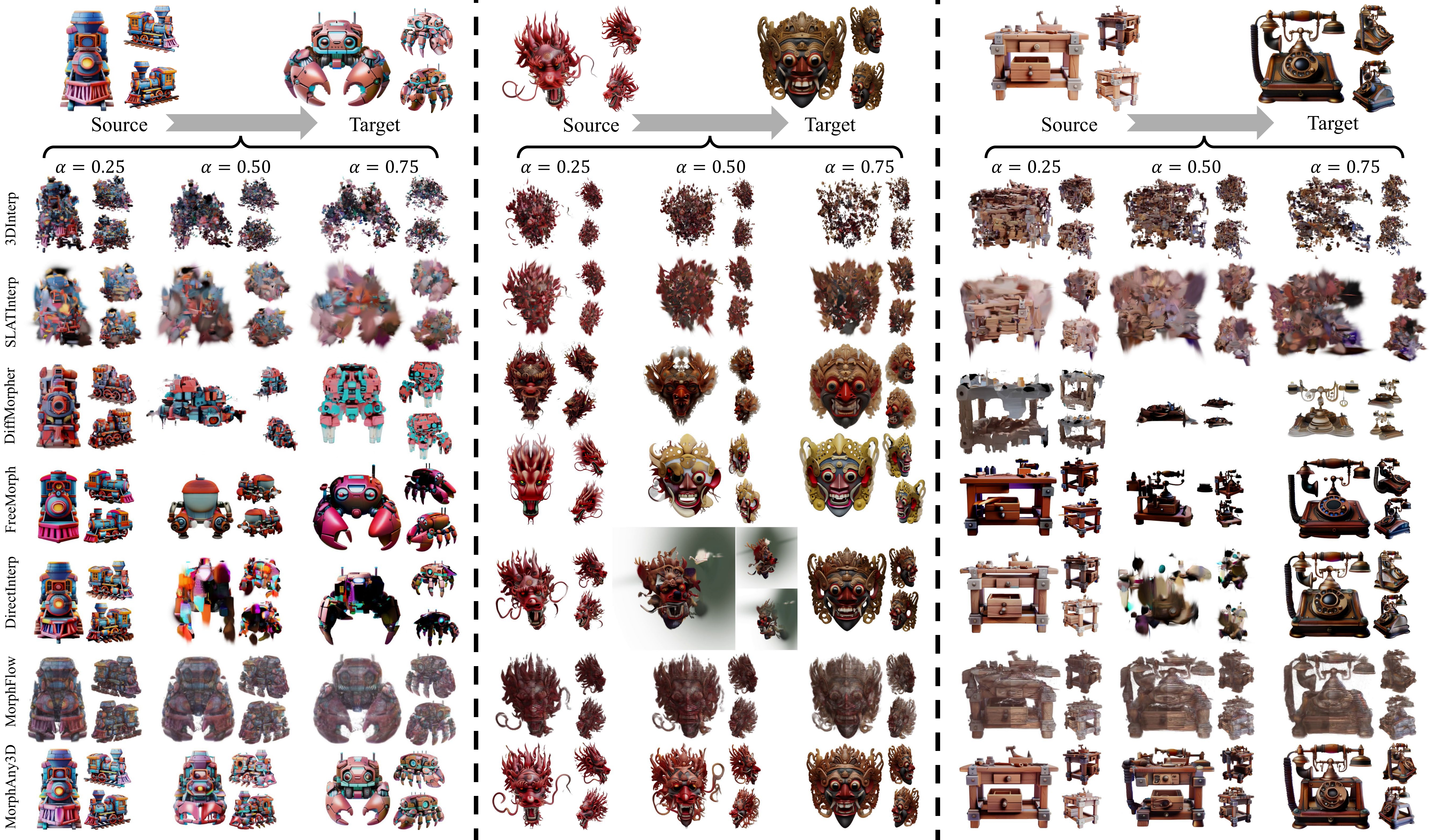} 
    \caption{Qualitative comparison with baseline methods. Our approach achieves superior morphing quality and smoother transitions.}
    \label{fig:comp1}
    \vspace{-0.3cm}
\end{figure*}

\begin{figure*}[tp]
    \centering
    \includegraphics[width=0.9\linewidth]{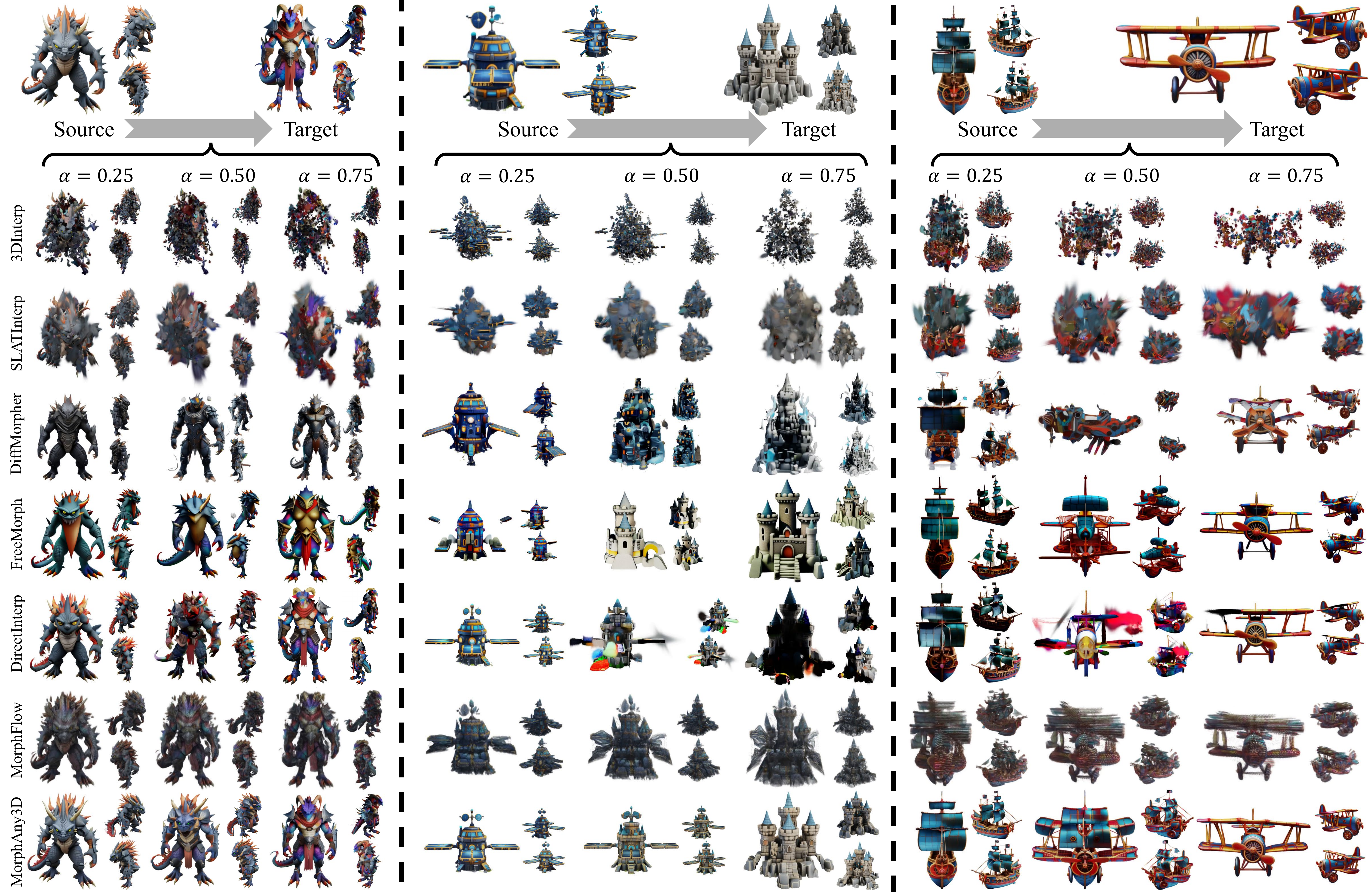} 
    \caption{Further qualitative comparisons highlighting the advantages of our method over existing approaches in preserving geometry and ensuring temporal coherence.}
    \label{fig:comp2}
    \vspace{-0.3cm}
\end{figure*}

Further comparisons appear in \cref{fig:comp1,fig:comp2}. 
Our method consistently outperforms baselines in morphing quality and temporal smoothness. 
Since 3DMorpher~\cite{3Dmorpher} has not released full code, we compare qualitatively using results reproduced from their paper (\cref{fig:comp3}).
We use the same source and target assets: cropped images from their figures are lifted to 3D via Image-to-3D Trellis~\cite{trellis}. 
As shown, our results exhibit superior detail and coherence.
Moreover, 3DMorpher struggles with complex structures (e.g., \cref{fig:show1,fig:show2}) and relies solely on 3DGS, which lacks high-fidelity normals and compatibility with standard 3D software~\cite{blender,maya}.

\section{Additional Application Details}
\textbf{Disentangled 3D Morphing.}
Thanks to Trellis's decoupled SS and SLAT stages, our method supports disentangled morphing. 
As illustrated in \cref{fig:app1}:
\begin{itemize}
    \item Structure-only morphing: Apply MCA/TFSA only in the SS stage to deform global shape while preserving local details.
    \item Detail-only morphing: Apply MCA/TFSA only in the SLAT stage to transfer texture and fine geometry without altering overall structure.
\end{itemize}

\noindent
\textbf{Dual-Target 3D Morphing.}
As shown in \cref{fig:app2}, distinct targets can be assigned to each stage: the SS stage adopts the source's global shape from one target, while the SLAT stage incorporates local details from another target.
This enables flexible composition of 3D assets, blending shapes and details from different sources.

\noindent
\textbf{3D Style Transfer.}
By conditioning the SLAT stage on a style reference while keeping the SS stage fixed to the source structure, our method performs 3D style transfer (\cref{fig:app3}), preserving shape while adopting the reference's geometric and textural style.

\section{Generalization Details}
Hi3DGen~\cite{hi3dgen}—a recent SLAT-based method for high-fidelity 3D generation—uses the same Trellis framework and image conditioning. Thus, our approach applies directly without retraining. Since Hi3DGen focuses on geometry (no texture), we report only normal maps.

For Text-to-3D Trellis~\cite{trellis}, we replace image conditions in MCA with textual prompts. No other changes are needed, enabling zero-shot adaptation. However, textual conditions are more abstract than visual ones, leading to reduced deformation quality and temporal coherence. We therefore recommend image-guided SLAT models when possible.

\end{document}